\documentclass{article}

\usepackage{microtype}
\usepackage{graphicx}
\usepackage{subcaption}
\usepackage{booktabs} 
\usepackage{stfloats}

\usepackage{hyperref}

\usepackage{amsmath}
\usepackage{amssymb}
\usepackage{mathtools}
\usepackage{amsthm}
\usepackage{stfloats}
\usepackage{url}
\usepackage{multirow}
\usepackage{multicol}
\usepackage{booktabs}
\usepackage[table]{xcolor}
\usepackage{enumitem}

\usepackage[preprint]{icml2026}

\usepackage[capitalize,noabbrev]{cleveref}

\theoremstyle{plain}

\theoremstyle{definition}

\theoremstyle{remark}

\usepackage[textsize=tiny]{todonotes}
\icmltitlerunning{~\ours{}: Hierarchical Speculative Decoding for LLMs}

\newcommand{\ignore}[1]{}
\newcommand{\ours}[0]{HiSpec}

\usepackage{tikz}

\newcommand{\BlueComment}[1]{\hfill{\color{blue}// #1}}
\newcommand{\AlgoComment}[1]{{\color{blue}// #1}}
\usepackage{xcolor}
\usepackage{tcolorbox}

\usepackage{pifont}
\definecolor{ForestGreen}{RGB}{34,139,34}
\newcommand{\cmark}{\textcolor{ForestGreen}{\ding{51}}}
\newcommand{\xmark}{\textcolor{red}{\ding{55}}}

\definecolor{mycolor}{HTML}{DDF4E7}
\usepackage{contour}
\usepackage[normalem]{ulem}

\contourlength{0.8pt}
\newcommand{\myuline}[1]{%
  \uline{\phantom{#1}}%
  \llap{\contour{white}{#1}}%
}

\begin{document}

\twocolumn[
  \icmltitle{~\ours{}: Hierarchical Speculative Decoding for LLMs}



  \icmlsetsymbol{equal}{*}

  \begin{icmlauthorlist}
    \icmlauthor{Avinash Kumar}{UT}
    \icmlauthor{Sujay Sanghavi}{UT}
    \icmlauthor{Poulami Das}{UT}
  \end{icmlauthorlist}

  \icmlaffiliation{UT}{Department of Electrical and Computer Engineering, The University of Texas at Austin}

  \icmlcorrespondingauthor{Avinash Kumar}{avinkumar@utexas.edu}

  \icmlkeywords{Machine Learning, ICML}

  \vskip 0.3in
]

\printAffiliationsAndNotice{}

\begin{abstract}

Speculative decoding accelerates LLM inference by using a smaller draft model to speculate tokens that a larger target model verifies. Verification is often the bottleneck (e.g. verification is $4\times$ slower than token generation when a 3B model speculates for a 70B target model), but most prior works focus only on accelerating drafting. \textit{``Intermediate"} verification reduces verification time by discarding inaccurate draft tokens early, but existing methods incur substantial training overheads in incorporating the intermediate verifier, increase the memory footprint to orchestrate the intermediate verification step, and compromise accuracy by relying on approximate heuristics.

We propose \textit{Hierarchical Speculative Decoding (\ours{})}, a framework for high-throughput speculative decoding that exploits \textit{early-exit (EE) models} for low-overhead intermediate verification. EE models allow tokens to exit early by skipping layer traversal and are explicitly trained so that hidden states at selected layers can be interpreted, making them uniquely suited for intermediate verification without drastically increasing compute and memory overheads.
To improve resource-efficiency even further, we design a methodology that enables \ours{} to re-use key-value caches and hidden states between the draft, intermediate verifier, and target models. To maintain accuracy, \ours{} periodically validates the draft tokens accepted by the intermediate verifier against the target model. Our results using various representative benchmarks and models show that \ours{} improves throughput by 1.28$\times$ on average and by up to 2.01$\times$ compared to the baseline single-layer speculation without compromising accuracy.

\end{abstract}
\section{Introduction}

Deploying Large Language Models (LLMs) presents inherent trade-offs between critical performance metrics, such as throughput, accuracy, and latency. Typically, larger models produce more accurate and coherent outputs but increase latency and reduce throughput. \textit{Speculative decoding} is an efficient inference technique that improves throughput by employing two models: a smaller \textit{draft} model that \textit{speculates} tokens and a larger \textit{target} model that verifies these tokens. Generating tokens using the draft model improves throughput, whereas verification against the target model ensures high accuracy comparable to the target model itself.

\begin{figure}[tp]
\begin{center}
    \centering
    \includegraphics[width=1.0\columnwidth]{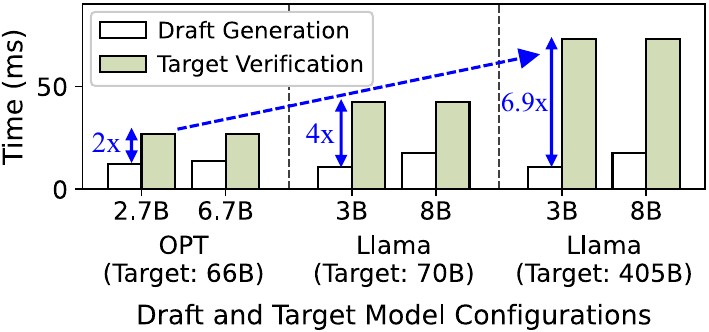}
    \caption{Token generation and verification latencies of different draft and target model combinations for ShareGPT dataset~\cite{sharegpt}. Verification takes $2$-$6.9\times$ longer than token generation and the gap between them grows with the size of the target models.}
\label{fig:verif_bottleneck}
\vspace{-0.1in}
\end{center}
\end{figure}

\begin{figure}[tp]
\begin{center}
    \centering
    \vspace{-0.1in}
    \includegraphics[width=1.0\columnwidth]{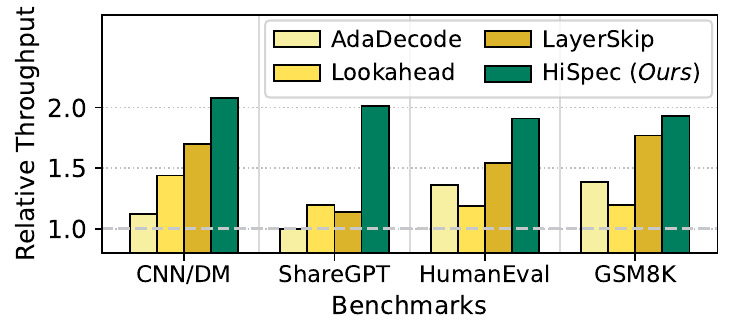}
    \vspace{-0.25in}
    \caption{Throughput of various representative benchmarks for Llama3-8B relative to auto-regressive decoding (\textit{higher is better}). By accelerating token verification, \ours{} consistently outperforms state-of-the-art prior works (AdaDecode, Lookahead Decoding, and LayerSkip) that accelerate draft token generation.}
\label{fig:results_summary}
\vspace{-0.2in}
\end{center}
\end{figure}

Token verification is significantly slower than draft generation because it involves processing more layers of a larger target model. Our studies with various configurations of draft and target models across model families show that verification takes $2-6.9\times$ longer than draft generation for the ShareGPT dataset with real-user conversations, as shown in Figure~\ref{fig:verif_bottleneck}. Slow verification severely limits the throughput of speculative decoding because subsequent tokens cannot be generated until previous tokens are verified due to the inherent sequential nature of token processing. The severity of this \textit{verification wall} scales with the size of target models. Unfortunately, most prior works accelerate draft token generation, resulting in only marginal throughput improvements, as shown in Figure~\ref{fig:results_summary}, while the verification bottleneck remains largely unaddressed.

Long token verification latencies can be addressed via ``\textit{intermediate}" verification. Prior work, 
SPRINTER~\cite{zhong2025speeding} uses an auxiliary model for intermediate verification and discards inaccurate tokens early. This also allows the next round of draft token generation to begin sooner. However, although SPRINTER improves throughput, it incurs significant computational and memory overheads to train the auxiliary model and orchestrate multiple models simultaneously. Moreover, it degrades accuracy because not all intermediate tokens are verified by the target model. \textit{Ideally, we want to enable low-overhead resource-efficient intermediate verification without sacrificing accuracy. }

\begin{figure}[tp]
\begin{center}
    \centering
    \includegraphics[width=1.0\columnwidth]{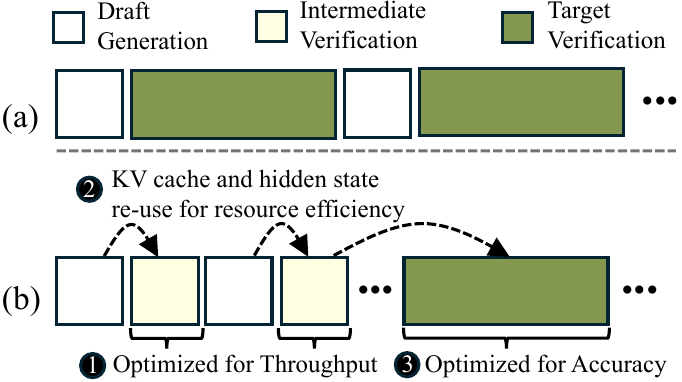}
    \caption{(a) In speculative decoding, a round of draft generation starts only after the previous (slow) verification round completes. (b)~\ours{} accelerates verification by using an intermediate verifier (thus improving overall throughput), re-uses KV caches and hidden states (thus improving resource-efficiency), and periodically invokes the target model verification (to optimize for accuracy).}
\label{fig:dummy_design}
\vspace{-0.25in}
\end{center}
\end{figure}

\textbf{Our Proposal:} We propose~\textit{\myuline{Hi}erachical \myuline{Spec}ulative Decoding (\ours{})}, a framework for high-throughput speculative decoding without degrading accuracy through low-overhead intermediate verification. \ours{} uses \textit{early-exit (EE)} models that allow tokens to skip traversing through an entire model by exiting early at selected layers and are explicitly trained so that hidden states at these exit layers can be interpreted.~\ours{} exploits this feature to employ early-exit layers for both draft token generation and intermediate verification, thereby
overcoming the computational overheads associated with the latter in prior work. To further reduce compute and memory overheads, we propose mechanisms to efficiently re-use Key-Value (KV) caches and hidden states across the draft, intermediate verifier, and target layers. To maintain output consistency with the target model, \ours{} also invokes periodic target (full-model) verification. Figure~\ref{fig:dummy_design} gives an overview of \ours{}.

Selecting an appropriate intermediate verifier is critical for overall performance. Using only a few model layers for intermediate verification reduces token acceptance rates by the target, whereas using too many layers reduces throughput. Our studies show that the earlier model layers (up to one-fourth the depth) are critical and generates up to 69\% of the response correctly. We use this insight to appropriately position the intermediate verifier that attains a sweet-spot in the throughput versus token acceptance rate trade-off space. 

Our evaluations show that \textit{\ours{} improves throughput by 1.28$\times$ on average and by up to 2.01$\times$ compared to the baseline single-layer speculation, without compromising accuracy.} \ours{} consistently outperforms state-of-the-art prior works that mainly accelerate draft token generation. We also show that~\ours{} is \textit{generalizable across both pre-trained and post-training modified EE-models} and, unlike prior work, \textit{natively supports batched inference}, facilitating seamless widespread adoption.

Overall, this paper makes the following contributions. 
\vspace{-0.1in}
\begin{enumerate}[leftmargin=2pt, itemindent=0pt, labelsep=2pt,itemsep=0pt]
    \item We show that long token verification latencies (upto {$6.9\times$} higher than draft generation) limits speculative decoding throughput because subsequent tokens cannot be speculated until prior tokens are verified.

    \item We propose \textit{Hierarchical Speculative Decoding (\ours{})} to addresses the verification bottleneck by using ~\textit{early-exit (EE)} models for \textit{low-overhead intermediate verification}. EE-models allow tokens to exit at specific layers, making them uniquely suited for intermediate verification without incurring computational overheads.
    
    \item We improve \ours{} resource-efficiency by re-using key-value caches and hidden states across the draft, intermediate verifier, and target to avoid redundant computations.
    
    \item We ensure that \ours{} maintains output consistency with what is otherwise produced by the target model, by enabling it to perform periodic target (full-model) verification.

\end{enumerate}
\begin{figure*}[t]
\begin{center}
    \centering
    \includegraphics[width=0.90\textwidth]{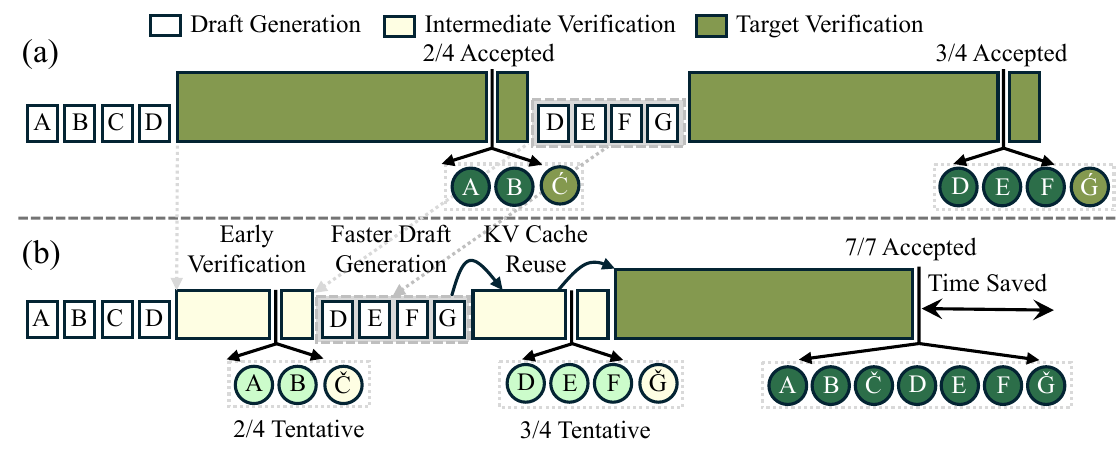}
    \caption{(a) Standard speculative decoding. (b) Our proposal, \textit{\ours{}}, uses \textit{early-exit models} for intermediate verification to reject inaccurate tokens early, thereby also accelerating subsequent draft generation. \ours{} reuses KV caches and hidden states to improve compute and memory efficiency and performs periodic target verification to maintain accuracy. }
    \vspace{-0.1in}
\label{fig:hispec_design}
\end{center}
\end{figure*}

\section{Background and Motivation}

\textbf{Speculative decoding:} Speculative decoding~\cite{leviathan2023fast, chen2023accelerating} is an efficient LLM inference technique that splits the inference process into two phases: draft generation and token verification. In the draft generation phase,
a smaller and faster but less accurate draft model speculates multiple tokens sequentially based on the preceding context. This is followed by token verification, in which a larger and more accurate target model verifies these predictions by evaluating the likelihood of each draft token against its own distribution. If a draft token does not match the token predicted by the target model, it is discarded along with all subsequent tokens. Otherwise, it is accepted. Note that a round of draft generation begins only after all draft tokens from the previous round have been verified.
Speculative decoding offers higher throughput than auto-regressive decoding while ensuring that the output is identical to that produced by the target model auto-regressively in isolation.

\textbf{Problem: \textit{Verification wall} limits performance:}
Token verification generally takes significantly longer than draft generation because it involves traversing larger models with complex architectures and more layers. Our studies with various Llama and OPT models (ranging between 1B to 405B) show that verification takes $2$-$6.9\times$ longer than draft generation and accounts for~60-90\% of the total response generation time, as shown in Figure~\ref{fig:verif_bottleneck} (please see Appendix~\ref{app:verif_overheads} for more details on our study). We refer to this bottleneck as the \textit{verification wall}. Most importantly, the severity of the verification wall scales with target model sizes and is expected to become even more critical in the future, as we adopt larger models to improve response quality.

\textbf{Limitations of prior works:}
Several prior works improve speculative decoding performance (discussed in detail in Appendix~\ref{app:related} on Related Work). However, these approaches mainly focus on optimizing token generation latency and improving acceptance rates (percentage of draft tokens accepted by the target model). Consequently, they yield limited returns because they do not reduce verification latencies that dictate overall runtime and throughput. 

In a recent work, Zhong et al. propose SPRINTER~\cite{zhong2025speeding} that speeds up token verification via \textit{intermediate verification} using an auxiliary model (which is larger than the draft but smaller than the target model). SPRINTER (1)~trains the auxiliary model to mimic the target model using approximate heuristics, (2)~employs it to \textit{discard inaccurate tokens early}, and (3)~invokes the target model only when absolutely necessary. By saving redundant computations for the early rejected tokens and minimizing the usage of the target model, SPRINTER improves throughput by up to $1.83\times$, emphasizing the benefits of intermediate verification. However, this approach incurs substantial overheads and degrades accuracy. This is because it requires computational resources to train the intermediate verifier. Moreover, SPRINTER degrades accuracy as the target model does not validate all tokens accepted by the intermediate verifier. 

\begin{tcolorbox}[
    colframe=blue!100,
    colback=white,
    colupper=blue!100,
    boxrule=0.6pt,
    arc=3mm,
]
\textbf{Goal-} Our goal is to accelerate token verification by enabling \textit{low-overhead and accurate intermediate verification without compromising accuracy}. 
\end{tcolorbox}
\section{Hierarchical Speculative Decoding~(\ours{})}

In this paper, we propose \textit{\myuline{Hi}erarchical \myuline{Spec}ulative Decoding (\ours{})}, a framework for high-throughput speculative decoding 
that leverages \textit{early-exit (EE) models} to enable low-overhead intermediate verification. EE-models are a class of LLMs designed to allow tokens to terminate layer traversal and exit at designated exit layer. This is achieved through explicit training to allow hidden states at these selected exit layers to be directly interpreted.~\ours{} exploits this feature to use these early exit layers for intermediate verification, thus eliminating any computational and memory overheads associated with integrating an auxiliary intermediate verifier.~\ours{} also employs an exit layer (lower than the intermediate verifier) for draft token generation. Thus, the draft and the intermediate verifier corresponds to early exit layers of the EE-model and the final layer (or full model) corresponds to the target. Figure~\ref{fig:hispec_design} gives an overview of \ours{}. Here, the first round of intermediate verification \textit{tentatively accepts} tokens $A$ and $B$, and rejects token $C$ early. This early verification also naturally enables faster draft generation in the next round, producing tokens $D$, $E$, $F$, and $G$, earlier than in the case of traditional speculative decoding, as shown in the Figure. To produce outputs consistent with what the target model produces auto-regressively,~\ours{} also periodically verifies against the full model. We first formalize~\ours{} and then discuss the implementation.

\subsection{Mathematical Formulation}
\label{sec:math}

\textbf{Early-Exit Model:} Let $\mathcal{M}$ be an early-exit model with a total of $L_f$ layers.~\ours{} leverages three exit-layers in $\mathcal{M}$. We use a shallow exit-layer $L_d$ to generate draft tokens, a higher exit-layer $L_i$ (where $L_d{<}L_i{<}L_f$) to perform intermediate verification, and the final layer $L_f$ to perform periodic full-model verification. $\mathcal{M}$ is trained such that hidden states at any $L \in \{L_d, L_i, L_f\}$, when projected through a shared or auxiliary LM head, yield a meaningful token-distribution over vocabulary $\mathcal{V}$.

\textbf{Auto-Regressive (AR) Decoding:} Let $x_{1:t}:= (x_1, x_2, \ldots, x_t) \in \mathcal{V}^{t}$ denote a token sequence of length $t$, and let $p_{L_f}(\cdot \mid x_{1:t})$ denote the next-token distribution obtained at the final layer $L_f$ of $\mathcal{M}$. Auto-regressive decoding samples one token at a time by running a full forward pass through $\mathcal{M}$:
\[x_{t+1} \;\sim\; p_{L_f}(\cdot \mid x_{1:t}). \quad\text{(Vanilla generation)}\]
\textbf{Self-Speculative Decoding:} Traditional speculative decoding~\cite{leviathan2023fast} pairs a fast draft model with a more accurate target model. In contrast, self-speculative decoding leverages a shallow layer $L_d$ to generate draft tokens, and uses the final-layer $L_f$ to verify them. In each round, the draft generates $N_d$ tokens, which the target verifies by performing a single forward pass. The draft distribution over each generated token i.e., $p_{L_d}(\cdot\mid x_{1:t})$ is then compared against the target i.e., $p_{L_f}(\cdot\mid x_{1:t})$ using an $\arg\max$ (greedy) rule to accept a contiguous prefix of draft tokens. At the first mismatch, the rejected token is replaced with the target's $\arg\max$ prediction and the successive draft tokens are discarded. We provide an overview in Algorithm~\ref{alg:self-spec}.

\begin{algorithm}[htp]
\caption{Self-Speculative Decoding}
\label{alg:self-spec}
\begin{algorithmic}[1]
\STATE \textbf{Input:} Draft layer $L_d$, Full layer $L_f$, 
\STATE \hspace{2.8em} Input Sequence $x_{1:t}$,
\STATE \hspace{2.8em} Draft proposal length $N_d$
\WHILE{not end-of-sequence}
\STATE $\mathbf{S} \gets \textsc{Generate}(L_d,\ x_{1:t},\ N_d)$
\STATE \AlgoComment{Prefix accepted by $L_f$ + bonus token}
\STATE $\mathbf{F} \gets \textsc{LeadingSubstringVerify}(\mathbf{S},\ L_f,\ x_{1:t})$
\STATE $x_{1:t} \gets x_{1:t} \Vert \mathbf{F}$
\ENDWHILE
\STATE \textbf{return} $x_{1:t}$
\STATE \textbf{function}\textsc{ LeadingSubstringVerify}($\mathbf{S},\ L,\ x_{1:t}$)
\STATE $v_k{\gets}\arg\max_{v\in\mathcal{V}} p_L(v{\mid}x_{1:t}, \mathbf{S}_{1:k-1}),\; k{=}1,\dots,|\mathbf{S}|{+}1$
\STATE $k^{\star} \gets \min\{k \in [1,|\mathbf{S}|] : \mathbf{S}_k \neq v_k\}$, else $|\mathbf{S}|{+}1$
\STATE \textbf{return} $\big[\,\mathbf{S}_{1:k^{\star}-1},\ v_{k^{\star}}\,\big]$
\STATE \textbf{end function}
\end{algorithmic}
\end{algorithm}

\textbf{\ours{}:} Baseline self-speculative decoding incurs the computational cost of an entire forward pass through $\mathcal{M}$ for every $N_d$ draft tokens, regardless of their accuracy. Our \textbf{key insight} is that the intermediate layers of $\mathcal{M}$ are accurate enough to early-reject inaccurate tokens at a fraction of the cost (compared to $L_f$). Thus,~\ours{} uses an intermediate layer $L_i$ to early-reject draft tokens and periodically invokes $L_f$ for full verification once $N_i$ tokens have been \textit{tentatively accepted}. This amortizes the cost of full forward passes through $\mathcal{M}$ across $N_i$ tokens, reducing the overall verification overhead. We provide an overview in Algorithm~\ref{alg:hispec}.

\begin{algorithm}[!b]
\caption{Hierarchical Speculative Decoding (\ours{})}
\label{alg:hispec}
\begin{algorithmic}[1]
\STATE \textbf{Input:} Draft layer $L_d$, Intermediate layer $L_i$
\STATE \hspace{2.8em} Full layer $L_f$, Input Sequence $x_{1:t}$,
\STATE \hspace{2.8em} Draft proposal length $N_d$,
\STATE \hspace{2.8em} Tentative acceptance window $N_i$
\WHILE{not end-of-sequence}
\STATE $\mathbf{V} \gets [\,]$ \BlueComment{Tokens tentatively accepted by $L_i$}
\WHILE{$|\mathbf{V}| < N_i$ \textbf{ and } not end-of-sequence}
\STATE $\mathbf{S} \gets \textsc{Generate}(L_d,\ x_{1:t}\Vert\mathbf{V},\ N_d)$
\STATE \AlgoComment{Prefix accepted by $L_i$ + bonus token}
\STATE $\widehat{\mathbf{V}} \gets \textsc{LeadingSubstringVerify}(\mathbf{S},\ L_i,\ x_{1:t}\Vert\mathbf{V})$
\STATE $\mathbf{V} \gets \mathbf{V}\Vert\widehat{\mathbf{V}}$
\ENDWHILE
\STATE \AlgoComment{Prefix accepted by $L_f$ + bonus token}
\STATE $\mathbf{F} \gets \textsc{LeadingSubstringVerify}(\mathbf{V},\ L_f,\ x_{1:t})$
\STATE $x_{1:t} \gets x_{1:t}\Vert\mathbf{F}$
\ENDWHILE
\STATE \textbf{return} $x_{1:t}$
\STATE \textbf{function}\textsc{ LeadingSubstringVerify}($\mathbf{S},\ L,\ x_{1:t}$)
\STATE $v_k{\gets}\arg\max_{v\in\mathcal{V}} p_L(v{\mid}x_{1:t}, \mathbf{S}_{1:k-1}),\; k{=}1,\dots,|\mathbf{S}|{+}1$
\STATE $k^{\star} \gets \min\{k \in [1,|\mathbf{S}|] : \mathbf{S}_k \neq v_k\}$, else $|\mathbf{S}|{+}1$
\STATE \textbf{return} $\big[\,\mathbf{S}_{1:k^{\star}-1},\ v_{k^{\star}}\,\big]$
\STATE \textbf{end function}
\end{algorithmic}
\end{algorithm}

\textbf{Lossless Guarantee:} Each token accepted by~\ours{} matches the $\arg\max$ of $L_f$ at that position. This is because $L_i$ only early-rejects inaccurate draft tokens, it does not relax the final acceptance  criterion. Thus, by successive application of the lossless speculative decoding proof~\cite{leviathan2023fast} at both verification stages i.e., $L_d{\to}L_i$ and $L_i{\to}L_f$,~\ours{} ensures outputs remain identical to AR decoding. Next, we discuss the design choices in~\ours{}.

\subsection{Dynamic KV and Hidden States Management For Low-Overhead Intermediate Verification}
\label{sec:kv_management}

Employing early exits for intermediate verification eliminates any additional training overheads. However, it alone is insufficient to reduce computational overheads because the forward pass for the generation of draft tokens and both intermediate and full-model verification still incurs redundant computations on the same input. To address this issue, \ours{} builds mechanisms to re-use the Key-Value (KV) caches and hidden states across the draft, intermediate verifier, and the target (full model). This is non-trivial because it requires careful alignment of these data structures. Even the slightest misalignment leads to a cascade of errors, which propagates through subsequent layers and token generation steps, ultimately corrupting the generated output. 

To overcome this challenge, we take a two-step approach. \textit{First}, during token generation, the generated KV caches and the hidden states are buffered separately until intermediate verification. \textit{Next}, we discard the KV pairs and hidden states associated with the rejected tokens to ensure that the subsequent phase begins with the correct context.~\ours{} manages the KV caches and hidden states by expanding them during the draft generation phase and pruning them back after intermediate verification to eliminate entries associated with tokens deemed inaccurate. This approach differs fundamentally from prior single-layer based speculation methods, where these structures are not repeatedly buffered or managed across arbitrary intermediate layers. During target verification, the accumulated hidden states are consumed to compute the acceptance outcome at the final layer, followed by elimination of inaccurate tokens from the KV cache before the next round of draft generation.

\subsection{Positioning Intermediate Verifier Layer}

The positioning of the intermediate verifier layer is critical to maximize the performance of \ours{}. For example, intuitively, using an intermediate verifier layer much higher than the draft layer (\textit{closer to the target}) would improve token acceptance rates, as deeper layers are more accurate and propagate only high-accuracy tokens to the final layer. However, this also increases the latency of intermediate verification and reduces throughput. In contrast, selecting an intermediate verifier at a much lower layer (\textit{closer to the draft}), yields high throughput but degrades token acceptance rates because it cannot adequately discard inaccurate tokens.

\begin{figure}[tp]
    \begin{center}
    \includegraphics[width=\columnwidth]{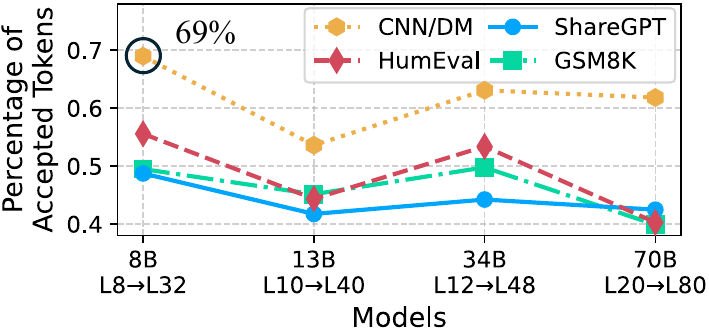}
    \vspace{-0.20in}
    \caption{Percentage of tokens produced by $1/4^{\text{th}}$ the model that are accepted by the final layer across diverse tasks for Llama models. Each label denotes one-fourth the model and its final layer (such as $L8 \rightarrow L32$ for Llama-8B with 32 layers). We observe that about $1/4^{\text{th}}$ of the model is sufficient to generate up to 69\% of the output tokens correctly. We use this information to position the intermediate verifier in \ours{}.}
    \label{fig:early}
    \vspace{-0.30in}        
    \end{center}
\end{figure}

To position the intermediate layer appropriately, we conduct studies using various models and observe that the early layers are critical and about one-fourth of the model layers generate up to 69\% of the response correctly, as shown in Figure~\ref{fig:early}. Moreover, prior work~\cite{jawahar2019does} shows that shallow semantic features, which capture high-level language-representations stabilize at about one-fourth the model depth. By building on our experimental results and similar observation from prior work, we position the intermediate verifier near one-fourth of the model depth, while allocating about one-eight of the layers for draft token generation in the default implementation of \ours{}.

\begin{figure}[!b]
\begin{center}
    \centering
    \vspace{-0.1in}
    \includegraphics[width=0.85\columnwidth]{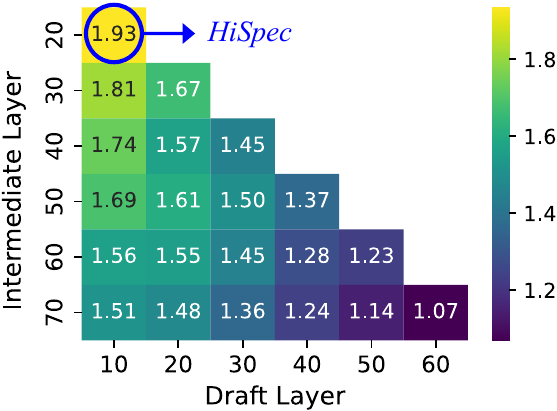}
    \caption{Throughput of \ours{} for different draft and intermediate layer choices relative to auto-regressive decoding for the CNN/DM dataset using Llama2-70B (80 layers). \ours{}'s selection of draft and intermediate verifier (circled) yields the highest throughput.}
\label{fig:heatmap}
\end{center}
\end{figure}

We conduct additional experiments to verify the efficacy of our intermediate verifier selection.  We compute the throughput of standard speculative decoding (baseline) using layer 10 for draft token generation and all 80 layers used for verification for the Llama2-70B model. The throughput is $1.47\times$ compared to vanilla auto-regressive decoding. Next, we use different exit layers for draft generation and intermediate verification in our study. Figure~\ref{fig:heatmap} shows the throughput for each draft and intermediate verifier combination relative to auto-regressive decoding. We observe that the default configuration of \ours{} achieves the maximum throughput of $1.93\times$. Nonetheless, even if \ours{} were to use another configuration, such as draft at layer 10 and intermediate verification at any layer ranging from 30 to 70, the throughput (between $1.51-1.81\times$) is still higher than the baseline.   

The default configuration of \ours{} offers the highest throughput because it achieves a sweet-spot in the throughput versus token acceptance rates trade-off space. Nevertheless, we emphasize that this default configuration is just intended as a practical guideline rather than a rigid architecture constraint for real-world deployment. We include additional details in Appendix~\ref{app:llama3_selection}, including the optimal intermediate verification layer across benchmarks for all Llama models used in our evaluations.

\begin{figure*}[!t]
    \centering
    \includegraphics[width=1.0\textwidth]{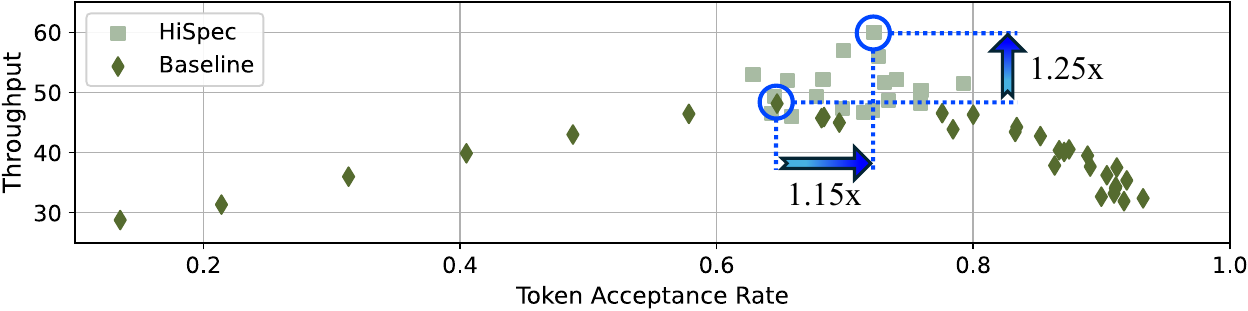}
    \vspace{0.05in}
    \caption{Comparison of token acceptance rates and throughput (\textit{higher is better for both}) for baseline single-layer early-exit speculation (Layerskip) against~\ours{} for CodeLlama-34B model (with 48 layers). Each datapoint for the baseline~\cite{elhoushi2024layerskip} corresponds to increasing number of layers used for draft token generation (47 possibilities). Each datapoint for \ours{} corresponds to the draft using exit Layer-7 (baseline optimal) and using increasing number of layers for intermediate verification (40 possibilities).~\ours{} improves both throughput and token acceptance rates.}
\label{fig:design}
\end{figure*}

\subsection{Balancing Intermediate and Full-Model Verification}
\label{sec:balance_intermediate}

Despite the advantages of intermediate verification, target verification is crucial to ensure that the generated output response is consistent with what is otherwise produced by the target model during with vanilla auto-regressive decoding. This is because the accuracy of the intermediate verifier is still lower than the target model and it is possible that a token \textit{tentatively accepted} by the intermediate verifier would be eventually rejected by the target. Target verification ensures that such tokens are discarded and the correct ones are regenerated. 
However, frequent target verification is compute-intensive, slow, and reduces throughput, whereas doing it too infrequently introduces high penalty even if one of the tokens in the tentatively accepted sequence is incorrect, because the entire sequence of tokens following this incorrect token (including itself) is flushed. The probability of this increases with the number of tentatively accepted tokens awaiting target verification.

To retain the accuracy benefits without lowering throughput, \ours{} performs multiple rounds of intermediate verification before invoking the target verification. However, naively invoking target verification every few rounds is sub-optimal because the number of tokens tentatively accepted by the intermediate verifier can vary.
Instead, \ours{} employs a dynamic policy that accumulates a sufficient number of tentatively accepted tokens before proceeding with target verification. By default, \ours{} waits until at least four tokens have been tentatively accepted (please see Section~\ref{sec:ablation_study} for more details).  
Nonetheless, the intermediate verifier leads to higher token acceptance rates than baseline single-layer speculation methods at the same draft layer. For example, our studies with the Llama2-70B model on the ShareGPT benchmark shows that the average token acceptance rate is only 39.7\% when using up to Layer-10 as the draft. \ours{} increases the average token acceptance rate to 58.1\% (46\% improvement) by employing an intermediate verifier that uses up to Layer-20.

Figure-\ref{fig:design} shows how \ours{} efficiently navigates the token acceptance rates and throughput. We consider the CodeLlama-34B model which comprises 48 layers. For the baseline, each datapoint corresponds to a draft with increasing number of layers and we consider all possible choices. Thus, we have 47 draft layer ($L_d$) choices, with a configuration using up to Layer-1, Layer-2, and so on up to Layer-47. As expected, relying on a higher layer for draft generation improves token acceptance rates. However, beyond a certain point ($L_d=7$), the overall throughput decreases because draft token generation takes too long. Figure-\ref{fig:design} also shows the performance for different \ours{} configurations, where each datapoint represents a draft and intermediate verifier combination. As using exit Layer-7 for the draft yields optimal performance in the baseline, we use the same in \ours{}. This leads to 40 possible combinations because now the intermediate verifier can use exit Layer-8, Layer-9, and so on up to Layer-47. We observe that the \ours{} configuration using exit Layer-7 and Layer-12 for the draft and intermediate verifier respectively improves throughput by $1.25\times$ and token acceptance rates by $1.15\times$. This is also consistent with the default \ours{} setting, which uses about one-fourth of the model for intermediate verification and one-eighth for draft token generation.
\begin{table*}[!b]
\centering
\vspace{-0.1in}
\caption{Qualitative comparison of~\ours{} against verification-acceleration works.~\ours{} is the only method that uses a single model, fully reuses KV cache, supports batched inference, is lossless, and requires no task-specific fine-tuning.}
\label{tab:verif-comp}
\setlength{\tabcolsep}{4pt}
\begin{tabular}{@{}lccccc@{}}
\toprule
\textbf{Property} & \textbf{HVSD} & \textbf{HSDDW} & \textbf{HSD} & \textbf{SPRINTER} & \textbf{\ours{} \textit{(Ours)}} \\ \midrule
No Auxiliary Models & \xmark{} & \xmark{} & \xmark{} & \xmark{} & \cmark{} \\
Full KV Reuse & \xmark{} (partial) & \xmark{} & \xmark{} & \xmark{} & \cmark{} \\
Batched Inference & \xmark{} & \xmark{} & \xmark{} & \xmark{} & \cmark{} \\
Lossless & \xmark{} (lossy) & \cmark{} & \cmark{} & \xmark{} (lossy) & \cmark{} \\
No Per-Task Tuning & \xmark{} (Bayesian opt) & \cmark{} & \cmark{} & \xmark{} (training) & \cmark{} \\
\bottomrule
\end{tabular}
\end{table*}

\begin{table*}[!b]
\centering
\caption{Comparison of \ours{} against SPRINTER across benchmarks on Llama3-8B. Beyond the auxiliary verifier, SPRINTER also uses a Llama3.2-1B draft model. Accuracy is computed using Llama3.1-70B as the judge. All metrics are relative to vanilla AR decoding.}
\label{tab:sprinter-comp}
\setlength{\tabcolsep}{6pt}
\begin{tabular}{@{}llccc@{}}
\toprule
\textbf{Benchmark} & \textbf{Metric (vs.\ AR)} & \textbf{SPRINTER} & \textbf{\ours{} \textit{(Ours)}} & \textbf{\ours{} / SPRINTER} \\ \midrule
\multirow{2}{*}{ShareGPT} & Throughput & $1.20\times$ & $\mathbf{2.01\times}$ & $1.67\times$ \\
& Accuracy & $0.40\times$ & $\mathbf{1.00\times}$ & $2.50\times$ \\
\midrule
\multirow{2}{*}{CNN/DM} & Throughput & $1.30\times$ & $\mathbf{2.08\times}$ & $1.60\times$ \\
& Accuracy & $0.52\times$ & $\mathbf{1.00\times}$ & $1.92\times$ \\
\midrule
\multirow{2}{*}{HumanEval} & Throughput & $1.32\times$ & $\mathbf{1.91\times}$ & $1.45\times$ \\
& Accuracy & $0.42\times$ & $\mathbf{1.00\times}$ & $2.38\times$ \\
\bottomrule
\end{tabular}
\end{table*}

\section{Evaluation Methodology}
\label{sec:eval}

\textbf{Benchmarks:} We select a diverse set of tasks, including real-user conversations, text summarization, code generation, and mathematical reasoning. Our selection is consistent with prior works (more details in Appendix~\ref{app:benchmarks}).

\textbf{Setup:} We conduct our experiments on a node with $4\times$ NVIDIA H100 GPUs (94GB HBM3). We implement~\ours{} using HuggingFace transformers~\cite{wolf2020transformers}, mirroring configuration of various prior works on speculative decoding~\cite{elhoushi2024layerskip, wei2025adadecode}.

\textbf{Models:} We evaluate~\ours{} across eight state-of-the-art models from the Qwen-3~\cite{yang2025qwen3}, Llama3~\cite{llama3}, CodeLlama~\cite{codellama}, Llama2~\cite{llama2}, and OPT~\cite{opt} families. To demonstrate \textit{wider applicability}, we use both pre-trained and post-training modified \textit{early-exit} (EE) models. The pre-trained EE variants of Llama are publicly available on HuggingFace~\cite{elhoushi2024layerskip}, while for the post-training setting, we augment off-the-shelf Qwen-3 and OPT families with prediction heads and fine-tune them to enable early-exits (more details in Appendix~\ref{app:post-training}).

\textbf{Baseline:} We compare~\ours{} against both works which accelerate draft generation and those which accelerate token verification. We consider AdaDecode, LayerSkip, SWIFT, and LookAhead decoding for draft acceleration. For works which accelerate token verification, we consider SPRINTER, HVSD, HSDDW and HSD. We provide more details on each baseline and their implementation in Appendix~\ref{app:baseline} and Appendix~\ref{app:baseline-implementation} respectively.

\textbf{Metrics:} We report throughput relative to AR decoding. We do not include accuracy metrics because~\ours{} uses greedy ($\arg\max$) decoding, and thus its output matches AR decoding exactly (except in Table~\ref{tab:sprinter-comp}). We also evaluate~\ours{} under rejection sampling in Appendix~\ref{app:rejection}.

We provide a summary of related works in Appendix~\ref{app:related}. This broad selection of benchmarks, models, and prior works establishes strong baselines for evaluating~\ours{}.
\section{Results}
\label{sec:results}

\subsection{Comparison With Verification Acceleration Work}

We compare~\ours{} against four prior verification-acceleration methods. HVSD~\cite{xianhierarchical} reduces verification overhead by skipping layers during intermediate verification, but it requires per-task Bayesian optimization to construct the intermediate verifier. Furthermore, this verifier is not transferable across tasks, limiting real-world deployment. On the other hand, HSDDW~\cite{syu2025hierarchical} and HSD~\cite{mohri2025fast} stack one or more auxiliary draft models between the smallest draft and the target. Finally, SPRINTER~\cite{zhong2025speeding} trains an auxiliary intermediate verifier that completely bypasses target verification when it is confident, making it \textit{lossy}. As none of these methods are open-source, we provide a qualitative comparison in Table~\ref{tab:verif-comp}. However, SPRINTER's training recipe is described in the original paper, thus, we re-implement and additionally report a quantitative comparison in Table~\ref{tab:sprinter-comp}.~\ours{} achieves $2.01\times$ throughput compared to $1.2\times$ on SPRINTER while maintaining output consistency with AR decoding whereas SPRINTER degrades accuracy by $60\%$.

\begin{table*}[!t]
\centering
\caption{Throughput of~\ours{} and other approaches that accelerate draft token generation relative to vanilla AR decoding (\textit{higher is better}). The best performance in each case is highlighted in {\bf bold}\label{tab:main_result}.\\}
\vspace{-0.1in}
\begin{tabular}{llccccc}
\toprule
\multirow{2}{*}{\textbf{Model}} & \multirow{2}{*}{\textbf{Method}} & \multicolumn{5}{c}{\textbf{Speedup (vs. AR)}} \\
\cmidrule(lr){3-7}
& & ShareGPT & CNN/DM & HumanEval & GSM8K & XSum \\
\midrule
\multirow{4}{*}{Llama3-8B}
& AdaDecode & 1.00$\times$ & 1.12$\times$ & 1.36$\times$ & 1.39$\times$ & 1.06$\times$ \\
& LayerSkip & 1.14$\times$ & 1.70$\times$ & 1.54$\times$ & 1.77$\times$ & 1.31$\times$ \\
& LookAhead & 1.20$\times$ & 1.44$\times$ & 1.19$\times$ & 1.20$\times$ & 1.37$\times$ \\
\rowcolor{mycolor}
& \ours{} \textit{(Ours)} & {\bf 2.01$\times$} & {\bf 2.08$\times$} & {\bf 1.91$\times$} & {\bf 1.93$\times$} & {\bf 1.84$\times$} \\
\midrule
\multirow{3}{*}{Llama2-7B}
& LayerSkip & 1.31$\times$ & 1.93$\times$ & 1.71$\times$ & 1.69$\times$ & 1.43$\times$ \\
& LookAhead & 1.37$\times$ & 1.34$\times$ & 1.24$\times$ & 1.50$\times$ & 1.43$\times$ \\
\rowcolor{mycolor}
& \ours{} \textit{(Ours)} & {\bf 1.70$\times$} & {\bf 1.95$\times$} & {\bf 1.84$\times$} & {\bf 1.92$\times$} & {\bf 1.50$\times$} \\
\midrule
\multirow{4}{*}{Llama2-13B}
& LayerSkip & 1.21$\times$ & 1.43$\times$ & 1.51$\times$ & {\bf 1.76$\times$} & 1.29$\times$ \\
& LookAhead & 1.26$\times$ & 1.55$\times$ & 1.23$\times$ & 1.43$\times$ & 1.45$\times$ \\
& SWIFT & 1.02$\times$ & 1.03$\times$ & 1.05$\times$ & 1.01$\times$ & 0.97$\times$ \\
\rowcolor{mycolor}
& \ours{} \textit{(Ours)} & {\bf 1.62$\times$} & {\bf 1.57$\times$} & {\bf 1.52$\times$} & 1.65$\times$ & {\bf 1.55$\times$} \\
\midrule
\multirow{5}{*}{CodeLlama-34B}
& AdaDecode & 1.19$\times$ & 1.53$\times$ & 1.67$\times$ & 1.56$\times$ & 1.37$\times$ \\
& LayerSkip & 1.23$\times$ & 1.42$\times$ & 1.50$\times$ & 1.52$\times$ & 1.42$\times$ \\
& LookAhead & 1.31$\times$ & 1.50$\times$ & 1.18$\times$ & 1.32$\times$ & {\bf 1.49$\times$} \\
& SWIFT & 1.09$\times$ & 1.13$\times$ & 1.24$\times$ & 1.12$\times$ & 1.05$\times$ \\
\rowcolor{mycolor}
& \ours{} \textit{(Ours)} & {\bf 1.55$\times$} & {\bf 1.59$\times$} & {\bf 1.67$\times$} & {\bf 1.60$\times$} & 1.45$\times$ \\
\midrule
\multirow{3}{*}{Llama2-70B}
& LayerSkip & 1.30$\times$ & 1.47$\times$ & 1.32$\times$ & 1.48$\times$ & 1.29$\times$ \\
& SWIFT & 1.10$\times$ & 1.17$\times$ & 1.12$\times$ & 1.09$\times$ & 1.10$\times$ \\
\rowcolor{mycolor}
& \ours{} \textit{(Ours)} & {\bf 1.64$\times$} & {\bf 1.93$\times$} & {\bf 1.69$\times$} & {\bf 1.72$\times$} & {\bf 1.63$\times$} \\
\bottomrule
\end{tabular}
\vspace{-0.1in}
\end{table*}

\begin{table*}[!b]
\centering
\vspace{-0.15in}
\caption{Throughput of~\ours{} and LayerSkip relative to vanilla AR decoding (\textit{higher is better}).~\ours{} outperforms LayerSkip even in the post-training setting. The best performance in each case is highlighted in {\bf bold} \label{tab:post-training}}
\begin{tabular}{llccccc}
\toprule
\multicolumn{1}{c}{\multirow{2}{*}{\textbf{Model}}}
& \multicolumn{1}{c}{\multirow{2}{*}{\textbf{Method}}}
& \multicolumn{5}{c}{\textbf{Speedup (vs. AR)}} \\
\cmidrule(lr){3-7}
& & ShareGPT & CNN/DM & HumanEval & GSM8K & XSum \\
\midrule
\multirow{2}{*}{Qwen3-8B}
& LayerSkip & 1.13$\times$ & 1.12$\times$ & 1.14$\times$ & 1.13$\times$ & 1.12$\times$ \\
& \ours{} \textit{(Ours)} & \textbf{1.42$\times$} & \textbf{1.36$\times$} & \textbf{1.44$\times$} & \textbf{1.41$\times$} & \textbf{1.35$\times$} \\
\midrule
\multirow{2}{*}{OPT-6.7B}
& LayerSkip & 1.36$\times$ & 1.52$\times$ & 1.46$\times$ & 1.47$\times$ & 1.44$\times$ \\
& \ours{} \textit{(Ours)} & \textbf{1.89$\times$} & {\bf 1.96$\times$} & {\bf 1.70$\times$} & {\bf 1.70$\times$} & {\bf 1.82$\times$} \\
\bottomrule
\end{tabular}
\end{table*}

\subsection{Comparison Against Draft Acceleration Methods}

Table~\ref{tab:main_result} compares \ours{} against prior works which accelerate draft generation. \ours{} improves throughput by $1.28\times$ on average and up to $2.01\times$ compared to these methods, emphasizing the criticality and benefits of accelerating verification.
Note that~\ours{} outperforms both Lookahead decoding and SWIFT, \textit{neither of which employ early-exits}, highlighting its efficacy even against speculative decoding strategies that do not employ early exits by default. 

\subsection{Generalizability Beyond Pre-Trained EE-Models}
\ours{} heavily exploits early-layers in EE models to improve throughput and can be integrated seamlessly with any EE model with at least two exit layers. Pre-trained EE models usually provide more exit layers, giving more flexibility to \ours{} for positioning the draft and intermediate verifier. For example, Llama checkpoints hosted by Layerskip allow tokens to exit at any arbitrary layer. Unlike pre-trained EE models, post-training modified EE models contain significantly fewer exit layers. This naturally constrains \ours{} by limiting the number of potential draft and intermediate verifier configurations. We evaluate~\ours{} for post-training modified EE-models by enabling early-exit capability in two off-the-shelf models (more details in Appendix~\ref{app:post-training}) from the Qwen-3 and OPT families. Table~\ref{tab:post-training} shows that \ours{} outperforms Layerskip even for these settings, despite limited flexibility. We choose Layerskip because it integrates seamlessly with any EE-model without incurring any additional training overheads. Thus, \textit{\ours{} remains effective for both pre-trained and post-training modified EE models. }

\begin{table*}[!h]
\centering
\caption{Throughput of~\ours{} and AR decoding for ShareGPT on Llama3-8B.~\ours{} improves throughput across batch sizes.}
\label{tab:batch_throughput}
\setlength{\tabcolsep}{6pt}
\begin{tabular}{@{}lccccc@{}}
\toprule
\textbf{Batch Size} & 1 & 2 & 4 & 8 & 16 \\ \midrule
Auto-regressive Decoding (tokens/sec) & 64.3 & 79.48 & 192.1 & 364.0 & 650.0 \\
\ours{} (tokens/sec) & \textbf{125.38} & \textbf{154.21} & \textbf{309.28} & \textbf{518.3} & \textbf{872.1} \\
Speedup & $1.95\times$ & $1.94\times$ & $1.61\times$ & $1.42\times$ & $1.34\times$ \\
\bottomrule
\end{tabular}
\vspace{-0.05in}
\end{table*}

\subsection{~\ours{} Supports Batched Inference}
By default, our evaluations use a batch size of one to match prior work. The fundamental challenge in batching speculative decoding is that each sequence accepts a different number of tokens per round, while standard batch operations require \textit{uniform} tensor shapes. Consequently, sequences are forced to operate in lockstep, thus, the first mismatch in any sequence limits the number of tokens that can be accepted by all sequences. Allowing each sequence to accept tokens independently, however, results in non-uniform KV cache lengths across the batch. Standard batched attention assumes a uniform KV layout, and therefore pads all other sequences to match the longest sequence. This introduces additional compute for each sequence, inflating per-token latency and offsetting the benefits of per-sequence acceptance. Consequently, \textit{prior work does not support batched inference}.~\ours{} solves these challenges by building mechanisms to incorporate paged-attention with block-tables, which computes attention per-sequence using only its valid KV blocks. This eliminates the computational overhead of padding and enables per-sequence acceptance without inflating per-token latency. Table~\ref{tab:batch_throughput} shows that~\ours{} consistently improves throughput across batch sizes.
\section{Ablation Study}
\label{sec:ablation_study}
\ours{} comprises three design parameters-- the intermediate verifier layer itself ($L_i$), the number of tokens produced by the draft layer per step ($N_d$), and the number of tokens tentatively accepted by the intermediate verifier ($N_i$). We analyze the impact of these parameters on \ours{}'s throughput using the ShareGPT dataset and Llama3-8B model. 

$\triangleright$  \textbf{Intermediate verifier layer ($L_i$):} 
As already discussed in the design Section~\ref{sec:balance_intermediate}, setting one-fourth of the model as the intermediate verifier maximizes throughput (and more details in Appendix~\ref{app:llama3_selection}).  We use this default configuration to study the impact of the other two parameters ($N_d$ and $N_i$).

\begin{figure}[htp]
    \includegraphics[width=1.0\columnwidth]{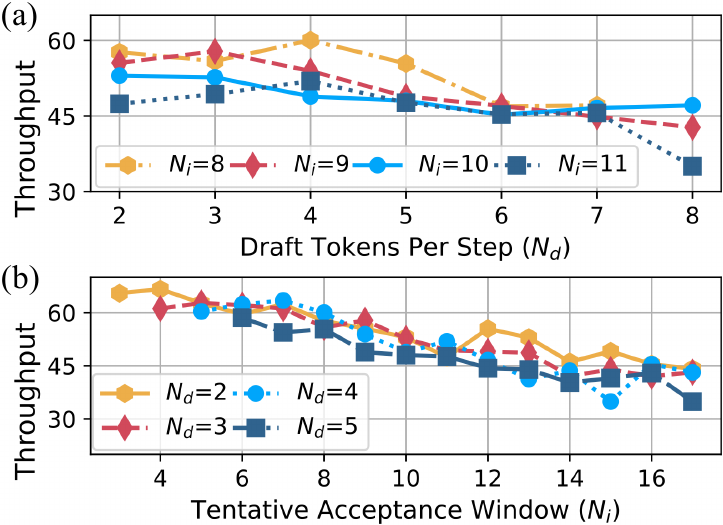}
    \vspace{-0.20in}
    \caption{Throughput with increasing (a) number of draft tokens per step ($N_d$) and (b) number of tokens tentatively accepted ($N_i$). Lower $N_d$ and $N_i$ yield higher throughput, which is expected, because they limit the formation of long chains of unverified tokens.}
    \vspace{-0.25in}
\label{fig:ni_variation}
\end{figure}

$\triangleright$  \textbf{Impact of number of draft tokens per step ($N_d$):} 
Too many draft tokens per step often increases the number of tokens flushed in case an incorrect token is identified. It also increases the latency of intermediate verification as a larger batch of draft tokens must be verified in parallel. Figure~\ref{fig:ni_variation}(a) shows that the throughput decreases with increasing $N_d$, which is consistent with our expectation. The default implementation of \ours{} uses $N_d=2$.

$\triangleright$  \textbf{Impact of number of tokens tentatively accepted ($N_i$):} Similarly, too many tentatively accepted tokens awaiting target verification also increase the number of tokens flushed and regenerated if the target model finds an inaccurate token. Figure~\ref{fig:ni_variation}(b) shows that the throughput decreases with increasing $N_i$, as we expect. By default, \ours{} uses $N_i=4$.

We also study an adaptive $N_d$ and $N_i$ selection policy that can be used to tune these parameters in real-time depending on the token acceptance rates to attain even higher throughput benefits. We present these details in Appendix-\ref{app:adapt-select}.
\section{Conclusion}

The throughput of speculative decoding is severely limited by the long verification latencies. 
We propose \textit{\myuline{Hi}erarchical \myuline{Spec}ulative Decoding (\ours{}}), a framework that addresses this bottleneck by using an intermediate verifier that \textit{tentatively accepts} draft tokens that it deems correct, and rejects them otherwise. Early rejection saves redundant computations, whereas tentatively accepted tokens accelerate the next draft generation round. To improve resource-efficiency, \ours{} proposes mechanisms to reuse key-value caches and hidden states across all stages. \ours{} also periodically invokes target verification to ensure output consistency with the target model. \ours{} improves throughput by 1.28$\times$ on average and up to 2.01$\times$ compared to baseline single-layer speculation methods and generalizes across both pre-trained and post-training modified EE-models.
\section{Acknowledgements}

This research was conducted using computing resources on the Vista GPU Cluster through the Center for Generative AI (CGAI) and the Texas Advanced Computing Center (TACC) at the University of Texas at Austin. We thank the generous support from the Cockrell School of Engineering and the Amazon AI PhD Fellowship Program through the Amazon Science Hub at UT Austin, including the AWS cloud credits that enabled the large-scale experiments conducted in this work. Poulami Das acknowledges the generous support through the AMD endowment at UT Austin.

\bibliography{references}
\bibliographystyle{icml2026}

\newpage
\appendix
\onecolumn

\begin{center}
\Large\bfseries Table of Contents
\end{center}

\noindent\textbf{\ref{app:verif_overheads} \quad Evaluating the Criticality of Accelerating Verification} \dotfill
\pageref{app:verif_overheads}\\[0.2em]

\noindent\textbf{\ref{app:llama3_selection} \quad Impact of Intermediate Verifier Selection} \dotfill \pageref{app:llama3_selection}\\[0.2em]

\noindent\textbf{\ref{app:benchmarks} \quad Benchmarks Used for Evaluations} \dotfill \pageref{app:benchmarks}\\[0.2em]

\noindent\textbf{\ref{app:related} \quad Related Work} \dotfill \pageref{app:related}\\[0.2em]

\noindent\textbf{\ref{app:baseline} \quad Evaluation Baselines} \dotfill \pageref{app:baseline}\\
\quad \ref{app:baseline-draft} \quad Draft-Generation Acceleration Methods \dotfill \pageref{app:baseline-draft}\\
\quad \ref{app:baseline-verif} \quad Token-Verification Acceleration Methods \dotfill \pageref{app:baseline-verif}\\[0.2em]

\noindent\textbf{\ref{app:baseline-implementation} \quad Baseline Implementation} \dotfill \pageref{app:baseline-implementation}\\[0.2em]

\noindent\textbf{\ref{app:rejection} \quad Performance of~\ours{} with Rejection Sampling} \dotfill \pageref{app:rejection}\\
\quad \ref{app:rejection-rule} \quad Acceptance Rule in Rejection-Sampling \dotfill \pageref{app:rejection-rule}\\
\quad \ref{app:rejection-lossless} \quad Lossless Guarantee under Rejection Sampling \dotfill \pageref{app:rejection-lossless}\\
\quad \ref{app:rejection-empirical} \quad Empirical Comparison \dotfill \pageref{app:rejection-empirical}\\[0.2em]

\noindent\textbf{\ref{app:acceptance-length} \quad Utility of Intermediate Verification} \dotfill
\pageref{app:acceptance-length}\\[0.2em]

\noindent\textbf{\ref{app:post-training} \quad Procedure and Overheads to Enable Early-Exits Post-Training} \dotfill \pageref{app:post-training}\\[0.2em]

\noindent\textbf{\ref{app:adapt-select} \quad Impact of Adaptive Hyper-parameter Selection} \dotfill \pageref{app:adapt-select}\\
\quad \ref{app:adapt-nd} \quad Impact on Number of Draft Tokens per Step ($N_d$) \dotfill \pageref{app:adapt-nd}\\
\quad \ref{app:adapt-ni} \quad Impact on Number of Tokens Tentatively Accepted ($N_i$) \dotfill \pageref{app:adapt-ni}\\[0.2em]

\noindent\textbf{\ref{app:limitations} \quad Future Work} \dotfill \pageref{app:limitations}

\newpage
\section{Evaluating the Criticality of Accelerating Verification}
\label{app:verif_overheads}

Prior works improves the performance of speculative decoding by accelerating draft generation and verification. We investigate the criticality of accelerating verification comparing its latency against token generation latency. To measure latency, we use the vLLM~\cite{vllm} inference engine and various combinations of draft and target models that are selected from prior work~\cite{yan2024decoding, lasby2025sd, galim2025draft}. The draft model generates six tokens per iteration. We conducts all these experiments on a node with eight NVIDIA H100 GPUs interconnected by NVlink, evaluated on the ShareGPT dataset. 
Our evaluations show that accelerating verification is even more critical because it is $2-10.3\times$ slower than token generation, as shown in Table~\ref{tab:verif_large}. Also, the gap between verification and draft generation latencies scale with the size of the target models. 


\begin{table*}[h]
\centering
\caption{Comparison between the latency of draft generation ($T_{\textrm{gen}}$) and the latency of target verification ($T_{\textrm{verif}}$)) in speculative decoding. These results show that verification time dominates the total execution time and is significantly slower. Moreover, this gap increases with increasing model sizes.\label{tab:verif_large}.\\} 

\begin{tabular}{cccccc}
\toprule
\multirow{2}{*}{Target Model} & \multirow{2}{*}{Draft Model} 
  & \multicolumn{2}{c}{Latency (ms)} & \multirow{2}{*}{$T_{\textrm{verif}}/T_{\textrm{gen}}$} \\
\cmidrule(lr){3-4}
 &  & Generation ($T_{\textrm{gen}}$) & Verification ($T_{\textrm{verif}}$) & \\
\midrule

\multirow{3}{*}{Facebook/OPT-66B} 
 & Facebook/OPT-1.3B & 9.14 & \multirow{3}{*}{26.83} & 2.9$\times$ \\
 & Facebook/OPT-2.7B & 12.23 &  & 2.2$\times$ \\
 & Facebook/OPT-6.7B & 13.48 &  & 2$\times$ \\
\midrule

\multirow{3}{*}{Llama-3.1-70B$_{\textsc{inst}}$} 
 & Llama-3.2-1B$_{\textsc{inst}}$ & 7.10 & \multirow{3}{*}{42.54} & 6$\times$ \\
 & Llama-3.2-3B$_{\textsc{inst}}$ & 10.58 &                      & 4$\times$ \\
 & Llama-3.1-8B$_{\textsc{inst}}$ & 17.42 &                      & 2.4$\times$ \\
\midrule

\multirow{3}{*}{Llama-3.1-405B$_{\textsc{inst}}$} 
 & Llama-3.2-1B$_{\textsc{inst}}$ & 7.10 & \multirow{3}{*}{72.97} & 10.3$\times$ \\
 & Llama-3.2-3B$_{\textsc{inst}}$ & 10.58 &                      & 6.9$\times$ \\
 & Llama-3.1-8B$_{\textsc{inst}}$ & 17.42 &                      & 4.2$\times$ \\
\bottomrule
\end{tabular}
\end{table*}

\section{Impact of Intermediate Verifier Selection}
\label{app:llama3_selection}

The selection of the intermediate verifier is critical to efficiently navigate the trade-offs between throughput versus acceptance rates. Selecting an intermediate verifier that is too small improves throughput but may not be able to accurately identify \textit{incorrect} draft tokens. In contrast, choosing
a large intermediate verifier enables a more accurate rejection of incorrect draft tokens, but reduces throughput, defeating its purpose altogether. To maximize benefits, we must select a sweet-spot on this trade-off curve. 

\begin{figure}[!t]
\begin{center}
\centerline{\includegraphics[width=1.0\textwidth]{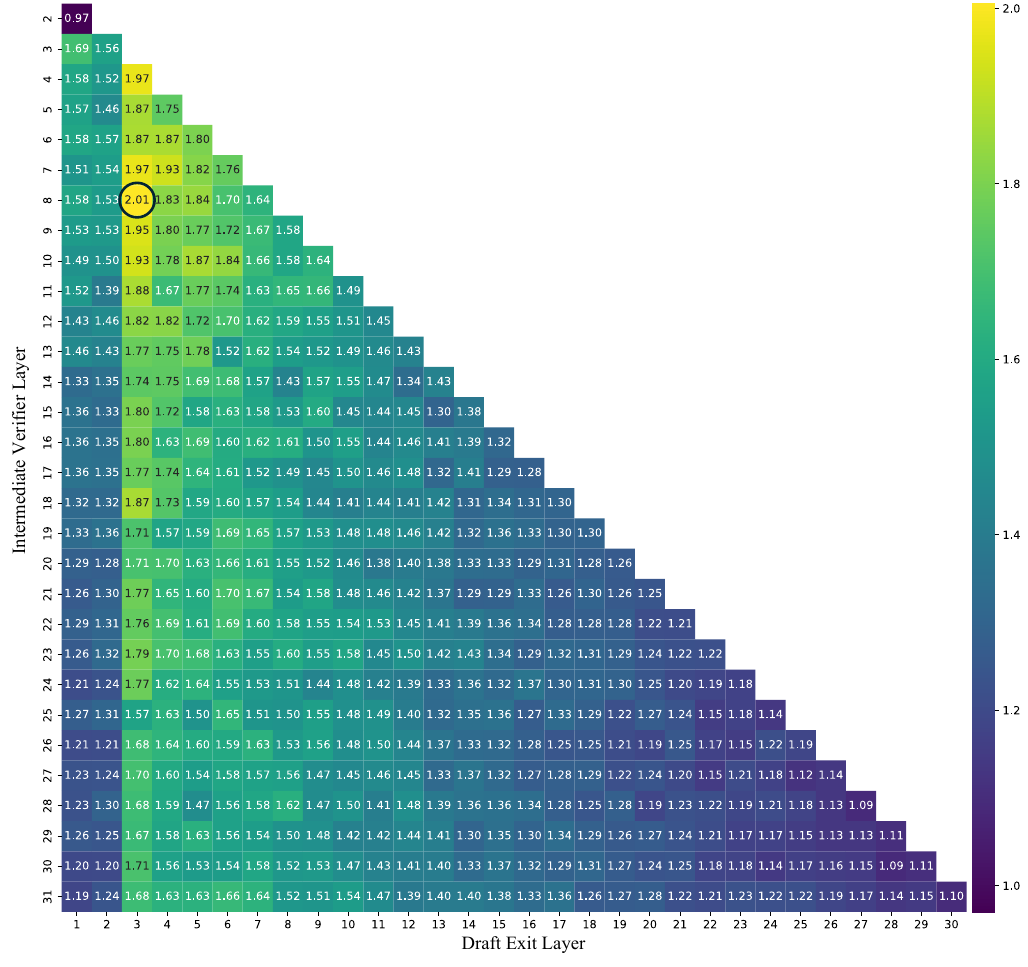}}
\caption{Throughput of 
all valid combinations of drafts and intermediate verifiers for the 32 layer Llama 3 8B model on the ShareGPT dataset relative to vanilla decoding (\textit{higher is better}). }
\label{fig:llama3-8B_finegrained}
\end{center}
\end{figure}

To assess the impact of intermediate verifier selection, we conduct an extensive study using the Llama 3.1 8B model as the target model and the ShareGPT dataset. It consists of 32 decoder layers. We evaluate different draft model size by varying the exit layers they use, sweeping it from Layer 1 to Layer 30. Similarly, we use different intermediate verifiers by sweeping the exit layers from Layer 2 to Layer 31. We analyze the throughput of all these draft and intermediate verifier combinations.
Figure~\ref{fig:llama3-8B_finegrained} shows the relative throughput of each combination compared to vanilla decoding for each draft and intermediate verifier combination.

Our evaluations show that among all configurations, the setting with draft generation at Layer 3 and intermediate verification at Layer 8 achieves the highest throughput. We use this setting as the default in our \ours{} design while using this model. In contrast, the configurations shown in the bottom row, where drafts are always verified only at Layer 31 degrades performance by up to 1.4$\times$. This setting approximates Layerskip based speculative decoding.
These results highlight that incorporating intermediate verification to discard inaccurate tokens \textit{early}, is critical for improving throughput. Moreover, it is important to select the right intermediate verifier to maximize throughput benefits.

Our experiments show that for the Layerskip~\cite{elhoushi2024layerskip}-hosted Llama checkpoints employed in our evaluations, the optimal intermediate verifier layer resides \textit{near} one-fourth the model depth as shown in Table~\ref{tab:interim_layer}. While~\ours{} builds on this observation to position the intermediate verifier at one-fourth the model depth by default,~\ours{} \textit{remains effective} even with alternate intermediate verifier layer configurations, as demonstrated in Table~\ref{tab:post-training}. 

\begin{table}[!h]
\centering
\caption{The optimal intermediate verifier layer in~\ours{} for across benchmarks for all Llama models used in our evaluations (Table~\ref{tab:main_result}).}
\begin{tabular}{lccccc}
\hline
\textbf{Model (Total Layers)} & \textbf{ShareGPT} & \textbf{CNN/DM} & \textbf{HumanEval} & \textbf{GSM8K} & \textbf{XSum} \\
\hline
Llama3-8B$_{\textsc{inst}}$ (32 Layers) & 8 & 10 & 8 & 8 & 9 \\
Llama2-7B$_{\textsc{inst}}$ (32 Layers) & 8 & 8 & 8 & 8 & 8 \\
Llama2-13B$_{\textsc{inst}}$ (40 Layers) & 10 & 15 & 12 & 10 & 10 \\
CodeLlama-7B$_{\textsc{inst}}$ (32 Layers) & 8 & 12 & 8 & 8 & 8 \\
CodeLlama-34B$_{\textsc{inst}}$ (48 Layers) & 12 & 16 & 12 & 12 & 12 \\
Llama2-70B$_{\textsc{inst}}$ (80 Layers) & 20 & 24 & 20 & 19 & 20 \\
\hline
\end{tabular}
\label{tab:interim_layer}
\end{table}

\clearpage
\section{Benchmarks Used for Evaluations}
\label{app:benchmarks}
We evaluate~\ours{} across a wide range of LLM tasks, including dialogue, text summarization, code-generation, and mathematical reasoning, covering a broad spectrum of model capabilities. The remainder of this section details the specific of the benchmarks used in this evaluation. Our selection of benchmarks is consistent with prior works. 

\textbf{Dialogue:} Drawn from real LLM interactions, ShareGPT~\cite{sharegpt} is an open-ended dialogue benchmark. The dataset was collected from users voluntarily sharing their GPT conversations, providing multi-turn exchanges. These prompts span a wide variety of topics, making the benchmark a realistic test of a model's ability to generate coherent, contextually appropriate responses.

\textbf{Text Summarization:} CNN Dailymail (CNN/DM)~\cite{cnn-dailymail} and Xsum~\cite{xsum} are popular summarization benchmarks. Both CNN/DM and Xsum are derived from online news articles. In CNN/DM, each article is paired with brief highlights summarizing the article. In contrast, Xsum provides a single-line summary of each article, emphasizing conciseness.

\textbf{Code Generation:} We use the HumanEval~\cite{humaneval} dataset to evaluate the robustness of~\ours{} on code-generation tasks. HumanEval consists of 164 python programming problems of varying difficulty. This benchmark suite assesses the ability of a model to produce functionally correct code.

\textbf{Mathematical Reasoning:} The GSM8K~\cite{gsm8k} benchmark suite contains over 8K grade school-level math word problems designed to assess arithmetic reasoning. Each prompt requires the model to perform step-by-step mathematical operations to arrive at the final solution.

\begin{figure}[H]
    \includegraphics[width=1.0\textwidth]{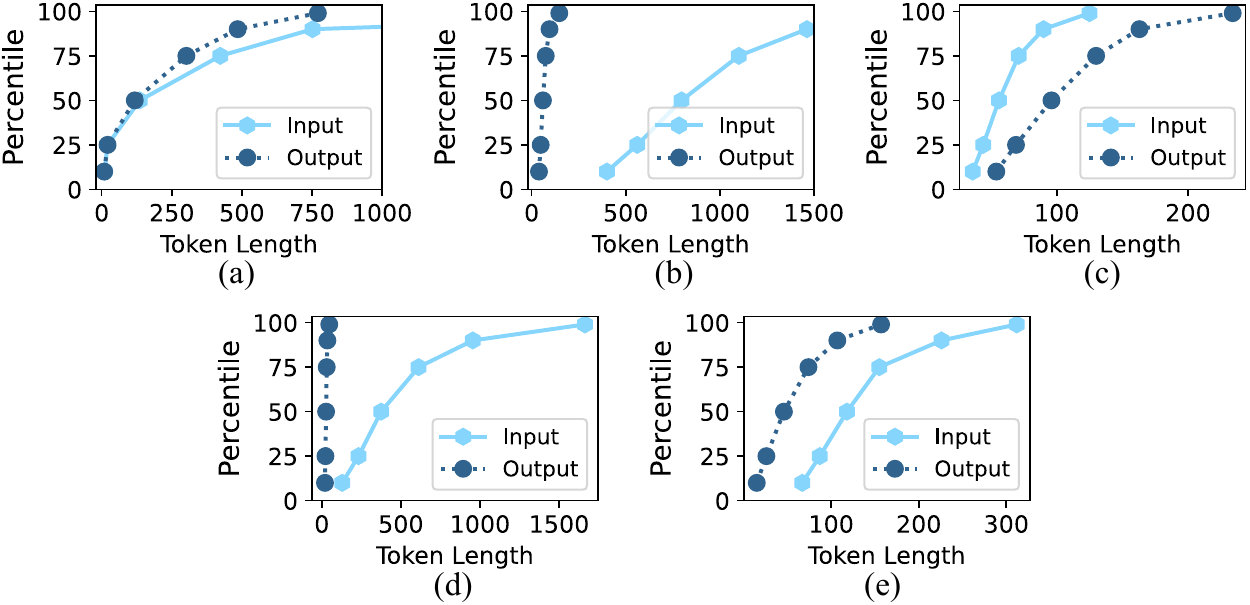}
    \caption{Cumulative distribution of input and output token lengths across (a) ShareGPT, (b) CNN/DM, (c) GSM8K, (d) Xsum, and (e) HumanEval. This distribution shows how the nature of the task affects token length. Summarization tasks (CNN/DM, Xsum) contain prompts with higher input token lengths. In contrast, mathematical reasoning tasks like GSM8K contain much shorter prompts.}
\label{fig:len_benchmark}
\end{figure}

\textbf{Input and Output Lengths:} Our selection of benchmarks exhibit substantial variation in both input and output token lengths, as shown in Figure~\ref{fig:len_benchmark}. For example, summarization tasks such as CNN/DM, Xsum contain longer input sequences compared to other benchmarks. These variations ensure that our evaluations encompasses a diverse range of input-output configurations.\\

\newpage
\section{Related Work}
\label{app:related}
Prior works improve the efficacy of speculative decoding by accelerating draft generation and target verification. Next, we present an overview of these directions, highlighting representative methods and their key limitations:

\textbf{Draft Generation:} Existing approaches that accelerate draft generation can be broadly categorized into methods which optimize the auxiliary draft model to generate accurate tokens or strategies which leverage the target model itself for draft generation. EAGLE~\cite{li2024eagle} designs a draft model to extrapolate the hidden representations of the target model, while SpecInfer~\cite{miao2023specinfer} explicitly optimizes the draft model to maximize token acceptance rates. However, these methods necessitate customized training and the reliance on an auxiliary draft model also introduces memory overheads~\cite{tang2025efficient} as the weights and KV caches of the auxiliary draft model need to be stored in GPU memory. 

In contrast, methods such as SWIFT~\cite{xia2024swift} selectively skip layers of the target model itself to efficiently generate draft tokens, but requires substantial workload-specific tuning to isolate layers and operations that can be skipped while maintaining high token acceptance rates. Lookahead decoding~\cite{fu2024break} modifies the attention mask to enable the target model to generate multiple tokens in parallel. However, this approach makes each decode step significantly more computationally expensive. Besides these approaches, methods like Layerskip~\cite{elhoushi2024layerskip}, and AdaDecode~\cite{wei2025adadecode} leverage \textit{early-exit models}, a class of language models which allows tokens to skip layer traversal at selected exit layers. Both these methods exploit this mechanism to employ an exit-layer within the target model for draft generation, and verify these predictions using the remaining target model layers. While Layerskip statically chooses a single draft layer, AdaDecode dynamically decides the draft layer by using confidence thresholds. Although these methods accelerate draft generation, overall throughput is still limited by the verification overhead, as demonstrated in Appendix~\ref{app:verif_overheads}.

\textit{These works mainly focus on accelerating draft token generation and cannot address the verification bottleneck, which is significantly more critical.}

\textbf{Target Verification:} To address the verification bottleneck, SPRINTER~\cite{zhong2025speeding} performs ``intermediate verification" using an auxiliary model that discards inaccurate tokens early and invokes the target model only when absolutely necessary. While this improves throughput, SPRINTER incurs substantial trade-offs. Orchestrating the draft, verifier, and the target model simultaneously increases the memory footprint as the weights and KV caches of all participating models need to be stored in GPU memory. Moreover, it introduces overheads to train a verifier model, capable of accurately approximate the target model. SPRINTER also degrades accuracy, as the target model does not validate all tokens accepted by the auxiliary intermediate verification model. 

Other recent works leverage hierarchical structures to accelerate verification but inherit similar trade-offs. HVSD~\cite{xianhierarchical} constructs a slim-verifier by skipping layers of the target and routes each draft token through it before invoking the full target. However, HVSD relies on tunable margin thresholds that relax the acceptance criterion, and thus, make it lossy. Moreover, constructing the slim verifier requires Bayesian optimization search for each benchmark, making it impractical for real-world deployments. HSDDW~\cite{syu2025hierarchical} and HSD~\cite{mohri2025fast} instead stack one or more auxiliary draft models between the smallest drafter and the target, where each auxiliary model serves as the intermediate verifier for the smaller drafter. While both these techniques are lossless, they incur substantial memory overheads from orchestrating multiple models concurrently. Moreover, none of these methods support batched inference.

Our proposal, \ours{}, addresses the verification bottleneck.~\ours{} exploits early-exit models to enable low-overhead intermediate verification using selected layers exit-layers. Unlike prior works that incur training overheads for intermediate verification and draft generation.~\ours{} can be naturally applied to any early-exit model (with at least two exits). \ours{} does not introduce any training or memory overheads and maintains the accuracy of the target model. Furthermore, unlike most prior work,~\ours{} natively supports batched inference.

\section{Evaluation Baselines}
\label{app:baseline}

We provide a description of each baseline used in our evaluation below. The baselines are organized into two categories based on whether they accelerate draft-generation or token verification.

\subsection{Draft-Generation Acceleration Methods}
\label{app:baseline-draft}

\textbf{AdaDecode}~\cite{wei2025adadecode} is the state-of-the-art early-exit speculation method. It dynamically selects a draft layer at runtime by evaluating the confidence of intermediate LM heads. However, AdaDecode requires task-specific fine-tuning of prediction heads for each benchmark. Furthermore, it also requires tuning a confidence threshold across each benchmark.

\textbf{LayerSkip}~\cite{elhoushi2024layerskip} statically selects a single exit-layer for draft generation and then verifies these tokens using the remaining layers of the same model. Similar to~\ours{}, Layerskip re-uses KV cache and hidden states across the draft generation and target verification stages, making it a strong baseline for comparison.

\textbf{SWIFT}~\cite{xia2024swift} selectively turns off model layers during draft generation. By exploiting redundancy across layers, SWIFT reduces compute for draft tokens on the fly, making it a strong baseline for comparing~\ours{} against layer-skipping methods. However, SWIFT requires per-task Bayesian search to determine the layer-skipping configuration.

\textbf{LookAhead Decoding}~\cite{fu2024break} modifies the attention mask to enable the target model to generate draft tokens in parallel, eliminating the need for a separate draft model. Unlike EE-based methods, it incurs more FLOPs per decode step.

\subsection{Token-Verification Acceleration Methods}
\label{app:baseline-verif}

All token-verification acceleration works use one or more auxiliary models along with the target. This imposes memory pressure and limits the memory available to support larger batch sizes or concurrently execute alternate applications.

\textbf{SPRINTER}~\cite{zhong2025speeding} trains a lightweight auxiliary verifier that predicts whether each draft token would be accepted by the target model. Tokens accepted by this verifier are committed without invoking the target model, this makes SPRINTER lossy since not all emitted tokens are validated by the target model.

\textbf{HVSD}~\cite{xianhierarchical} constructs a slim-verifier by skipping layers from the full target model, sharing weights with the target. HVSD conducts verification hierarchically, it directly emits high-confidence draft tokens, uses the slim-verifier to regenerate tokens which have medium-confidence and defer low-confidence tokens to the target. As not all tokens are verified by the target model, HVSD is \textit{lossy}. Moreover, HVSD requires per-task Bayesian optimization to determine the layer-skipping configuration of the slim-verifier, limiting its scalability across benchmarks.

\textbf{HSDDW}~\cite{syu2025hierarchical} orchestrates three models - a small primary draft, a relatively bigger draft and the much larger target model. The small draft model generates tokens, and the bigger draft model verifies these tokens. Only tokens accepted by the bigger draft model is propagated to the target for final verification. Each model maintains its own KV cache, with no re-use across stages.

\textbf{HSD}~\cite{mohri2025fast} stacks $K{+}1$ separate models in a chain, where each model acts as both drafter for the next-larger model and verifier for the smaller one. Standard rejection sampling is applied at each level, but memory overhead grows linearly with the number of stages because each model maintains its own KV cache.

\section{Baseline Implementation}
\label{app:baseline-implementation}

We describe below the implementation of the draft-acceleration works used for evaluations. Our baseline implementations are consistent with state-of-the-art prior work~\cite{wei2025adadecode}.

Both \textbf{Lookahead decoding}~\cite{fu2024break} and \textbf{Layerskip}~\cite{elhoushi2024layerskip} do not incur additional training overheads and work seamlessly with the EE-variants of Llama models used in our evaluations. Thus, we directly use their released code-base and \textit{perform an empirical search} to determine the optimal hyper-parameters across models and benchmarks.

\textbf{AdaDecode}~\cite{wei2025adadecode}: AdaDecode requires task-specific fine-tuning of prediction heads at early-exit layers. We use the training configuration proposed in their original implementation. We follow the training configuration from their original implementation and use the released task-specific checkpoints where available.

\textbf{SWIFT}~\cite{xia2024swift}: Following prior works~\cite{wei2025adadecode, zhang2024draft}, we perform a Bayesian search (200 iterations) to determine the layer-skipping configuration. Consistent with their original implementation, we randomly select four samples from each benchmark for this search. We choose an adaptive confidence threshold strategy and set the initial threshold as 0.6, the maximum number of draft token as 12, and the maximum number of generated token as 512.

\textbf{SPRINTER}~\cite{zhong2025speeding}: The code-base of SPRINTER is not open-source, thus, we re-implement it following the original paper. The intermediate verifier a single fully-connected layer with sigmoid activation ($2.1K$ parameters) that approximate whether the target model would accept or reject each token generated by the auxiliary draft model. Following the paper, we generate training labels by computing the probability ratio between the draft and the target model on prompts from the evaluation benchmarks. To remain consistent, we classify tokens with ratio below $1/\lambda$ ($\lambda{=}1.2$) as a mismatch.

\section{Performance of~\ours{} with Rejection Sampling}
\label{app:rejection}

The main paper presents~\ours{} with greedy ($\arg\max$) acceptance, which produces outputs identical to vanilla auto-regressive decoding from $p_{L_f}$ (Section~\ref{sec:math}). In this section, we additionally evaluate~\ours{} under the rejection sampling, another standard acceptance rule for speculative decoding~\cite{leviathan2023fast, chen2023accelerating}. We first show that~\ours{} preserves the target distribution under rejection-sampling by successive application of the rejection-sampling proof at both verification stages and then empirically compare the resulting throughput with the greedy acceptance. 

\subsection{Acceptance Rule in Rejection-Sampling}
\label{app:rejection-rule}

\textbf{Intermediate Verification ($L_d \to L_i$).} Given $N_d$ draft tokens $\mathbf{S}_1, \ldots, \mathbf{S}_{N_d}$ sampled from $p_{L_d}(\cdot \mid x_{1:t})$, and intermediate distributions $p_{L_i}(\cdot \mid x_{1:t}, \mathbf{S}_{1:k-1})$ at each position, for $k{=}1,..,N_d$ sequentially:
\begin{itemize}
    \item Draw $u \sim \text{Uniform}(0, 1)$.
    \item \textbf{Accept} $\mathbf{S}_k$ if $u < \min\left(1,\ \dfrac{p_{L_i}(\mathbf{S}_k \mid \cdot)}{p_{L_d}(\mathbf{S}_k \mid \cdot)}\right)$.
    \item \textbf{Reject} otherwise: discard $\mathbf{S}_k$ and all subsequent draft tokens, then sample a replacement from the residual distribution
    \[
    \mathbf{S}'_k \;\sim\; \frac{\max(0,\ p_{L_i}(v \mid \cdot) - p_{L_d}(v \mid \cdot))}{\sum_{v' \in \mathcal{V}} \max(0,\ p_{L_i}(v' \mid \cdot) - p_{L_d}(v' \mid \cdot))}.
    \]
\end{itemize}
The resulting buffer $\mathbf{V}$ includes the accepted draft tokens plus the replacement at the point of first mismatch sampled from the residual distribution. $\mathbf{V}$ is distributed as $p_{L_i}(\cdot \mid x_{1:t}, \mathbf{V}_{1:k-1})$ by the standard rejection-sampling lemma~\cite{leviathan2023fast}.

\textbf{Final Verification ($L_i \to L_f$).} Once $|\mathbf{V}| \ge N_i$, we apply the same rule with $\mathbf{V}$ as the draft and $p_{L_f}$ as the target. For $k = 1, \ldots, |\mathbf{V}|$ sequentially:
\begin{itemize}
    \item Draw $u \sim \text{Uniform}(0, 1)$.
    \item \textbf{Accept} $\mathbf{V}_k$ if $u < \min\left(1,\ \dfrac{p_{L_f}(\mathbf{V}_k \mid \cdot)}{p_{L_i}(\mathbf{V}_k \mid \cdot)}\right)$.
    \item \textbf{Reject} otherwise: discard $\mathbf{V}_k$ and all subsequent tokens, then sample a bonus token from
    \[
    \mathbf{F}^* \;\sim\; \frac{\max(0,\ p_{L_f}(v \mid \cdot) - p_{L_i}(v \mid \cdot))}{\sum_{v' \in \mathcal{V}} \max(0,\ p_{L_f}(v' \mid \cdot) - p_{L_i}(v' \mid \cdot))}.
    \]
\end{itemize}

\subsection{Lossless Guarantee under Rejection Sampling}
\label{app:rejection-lossless}

The intermediate verification stage transforms the draft distribution $p_{L_d}$ into $p_{L_i}$ by the standard rejection-sampling lemma. Every token in the buffer $\mathbf{V}$, whether it originates as an accepted draft or as a placement at the rejection point is distributed as $p_{L_i}(\cdot \mid x_{1:t}, \mathbf{V}_{1:k-1})$. The final verification stage applies the same lemma a second time, transforming $p_{L_i}$ into $p_{L_f}$. By composing the two stages, every emitted token in $\mathbf{F}$ is distributed as $p_{L_f}(\cdot \mid x_{1:t}, \mathbf{F}_{1:k-1})$, identical to vanilla auto-regressive sampling from $p_{L_f}$. This is a direct application of the standard speculative-decoding correctness proof~\cite{leviathan2023fast} applied sequentially at the two verification stages of \ours{}.

\subsection{Empirical Comparison}
\label{app:rejection-empirical}

Table~\ref{tab:rejection-throughput} compares the throughput of~\ours{} under greedy and rejection-sampling acceptance, both relative to AR decoding on the Llama3-8B ($L_d{=}4$, $L_i{=}6$) model for ShareGPT. Greedy is faster because it requires only a Top-1 comparison per token, whereas rejection sampling requires computing the full distributions $p_{L_d}, p_{L_i}, p_{L_f}$ and performing a stochastic accept/reject test at every position. Both, however, yield speedup over AR decoding while preserving the target distribution.

\begin{table}[ht]
\centering
\caption{Throughput of~\ours{} under greedy and rejection-sampling acceptance rules, relative to vanilla auto-regressive decoding on Llama3-8B for the same~\ours{} configuration i.e., $L_d{=}4$, $L_i{=}6$.}
\label{tab:rejection-throughput}
\setlength{\tabcolsep}{6pt}
\begin{tabular}{@{}lc@{}}
\toprule
\textbf{Acceptance Rule} & \textbf{Speedup vs.\ AR} \\
\midrule
Greedy ($\arg\max$, default) & $\mathbf{2.00\times}$ \\
Rejection sampling           & $\mathbf{1.54\times}$ \\
\bottomrule
\end{tabular}
\end{table}

\section{Utility of Intermediate Verification}
\label{app:acceptance-length}

\ours{}'s intermediate verifier $L_i$ tentatively accepts draft tokens, filtering out inaccurate predictions before they reach the final
$L_f$ verification. To concretely demonstrate this, we decompose intermediate verification into two metrics:

\textbf{True Reject Rate}: fraction of draft tokens correctly rejected by $L_i$ (also rejected by $L_f$). These tokens never incur the full-model verification cost.

\textbf{False Accept Rate}: fraction of draft tokens tentatively accepted by $L_i$ but later rejected by $L_f$. These tokens incur the full-model verification cost without contributing to acceptance.

Table~\ref{tab:acceptance-trade-off} reports these metrics, the resulting final acceptance rate, throughput, and peak memory overhead across
varying $L_i$ for Llama3-8B and Llama2-70B. As $L_i$ increases, the True Reject Rate rises and the False Accept Rate falls, since deeper
layers produce predictions that better align with $L_f$'s decisions. This translates to a higher final acceptance rate up to $95.39\%$ for
Llama3-8B at $L_i{=}30$, and up to $88.89\%$ for Llama2-70B at $L_i{=}76$, but at the cost of throughput, since deeper $L_i$ also makes
intermediate verification more expensive.

\begin{table}[ht]
\centering
\caption{Trade-off in intermediate verification across $L_i$ for Llama3-8B ($L_d{=}4$) and Llama2-70B ($L_d{=}8$) measured on a \textit{commercial} AMD 8xMI350x node. As $L_i$ increases, the True Reject Rate rises and the False Accept Rate falls, yielding a higher final acceptance rate but lower throughput.~\ours{} maintains ${\geq}1.32\times$ throughput across all $L_i$ with negligible memory overhead.}
\label{tab:acceptance-trade-off}
\setlength{\tabcolsep}{4pt}
\begin{tabular}{@{}llccccc@{}}
\toprule
\textbf{Model} & \textbf{$L_i$} & \textbf{True Reject} & \textbf{False Accept} & \textbf{Final Accept} & \textbf{Speedup vs.\ AR} &
\textbf{Mem. Overhead} \\ \midrule
\multirow{8}{*}{\shortstack{Llama3-8B \\ (32 layers)}}
 & 6  & $18.9\%$ & $14.5\%$ & $74.7\%$ & $2.23\times$ & $0.002\%$ \\
 & 8  & $23.6\%$ & $14.7\%$ & $73.8\%$ & $2.11\times$ & $0.002\%$ \\
 & 12 & $22.3\%$ & $11.4\%$ & $79.1\%$ & $2.00\times$ & $0.001\%$ \\
 & 16 & $22.4\%$ & $9.5\%$  & $82.6\%$ & $1.89\times$ & $0.004\%$ \\
 & 20 & $23.4\%$ & $7.8\%$  & $85.6\%$ & $1.43\times$ & $0.002\%$ \\
 & 24 & $24.2\%$ & $6.2\%$  & $89.4\%$ & $1.37\times$ & $0.005\%$ \\
 & 28 & $25.8\%$ & $4.8\%$  & $91.9\%$ & $1.32\times$ & $0.005\%$ \\
 & 30 & $22.9\%$ & $2.6\%$  & $95.4\%$ & $1.37\times$ & $0.003\%$ \\
\midrule
\multirow{8}{*}{\shortstack{Llama2-70B \\ (80 layers)}}
 & 12 & $62.9\%$ & $24.4\%$ & $55.8\%$ & $1.75\times$ & $0.001\%$ \\
 & 14 & $28.6\%$ & $21.8\%$ & $59.6\%$ & $1.75\times$ & $0.001\%$ \\
 & 20 & $35.3\%$ & $20.4\%$ & $62.8\%$ & $1.74\times$ & $0.001\%$ \\
 & 28 & $35.7\%$ & $17.8\%$ & $68.2\%$ & $1.55\times$ & $0.001\%$ \\
 & 44 & $34.1\%$ & $14.0\%$ & $74.5\%$ & $1.43\times$ & $0.001\%$ \\
 & 52 & $37.0\%$ & $10.9\%$ & $80.4\%$ & $1.39\times$ & $0.001\%$ \\
 & 70 & $36.7\%$ & $8.0\%$  & $86.6\%$ & $1.34\times$ & $0.002\%$ \\
 & 76 & $36.3\%$ & $6.7\%$  & $88.9\%$ & $1.32\times$ & $0.001\%$ \\
\bottomrule
\end{tabular}
\end{table}

We make two observations. \textit{First},~\ours{} consistently sustains a relative throughput of $1.32\times$ or higher across all $L_i$ choices,
confirming that even early intermediate rejection yields meaningful end-to-end speedup. \textit{Second}, the additional peak memory overhead
from intermediate verification is negligible ($<0.01\%$), since~\ours{} reuses the same KV cache and hidden-state buffer
across the draft, intermediate, and full-model stages (Section~\ref{sec:kv_management}). Thus, intermediate verification is a lightweight, resource-efficient mechanism that improves both throughput and acceptance length.

\section{Procedure and Overheads to Enable Early-Exits Post-Training}
\label{app:post-training}

For off-the-shelf models that do not natively support early-exits, prediction heads can be added at selected intermediate layers and fine-tuned via knowledge distillation against the final-layer output. This one-time process is significantly cheaper than training an auxiliary draft model from scratch. Moreover, post-training modified EE-models typically only train the prediction heads, while the rest of the model remains frozen. Thus, it does not affect the baseline accuracy of the target model. We describe the specific configurations used for the OPT and Qwen3 families below:

\textbf{OPT-6.7B:} We augment prediction heads at intervals of $1/4^{\text{th}}$ of the model depth and fine-tune these heads over 50K
 iterations on the Pile dataset~\cite{pile}. We use the first head (at $1/4^{\text{th}}$ depth) for draft generation and the second head (at
$1/2$ depth) for intermediate verification.

\textbf{Qwen3-8B.} For Qwen3-8B (36 total layers), we add low-rank prediction heads (rank 1024) at every fourth layer, i.e., at
layers $\{3, 7, 11, 15, 19, 23, 27, 31\}$. We train the heads via KL distillation against the final-layer output (temperature 2.0) on 100K
samples from ShareGPT, with batch size 4, learning rate $1{\times}10^{-4}$ on a cosine schedule with 100 warmup steps, for 16 epochs on a single GPU. We use Layer-3 for draft generation and Layer-7 for intermediate verification in~\ours{}.

\textbf{Constraint on~\ours{}.} Post-training modified EE-models contain significantly fewer exit layers than pre-trained models, which limits the $L_d$ and $L_i$ configurations in~\ours{}. Despite this
constraint, Table~\ref{tab:post-training} shows that~\ours{} outperforms LayerSkip, demonstrating that~\ours{} remains effective even when the EE-model offer much fewer exits.

\textbf{Training overhead compared to competitive methods:} We emphasize that early-exit capability is not a limitation specific to~\ours{}; rather,
 it is a fundamental requirement of \emph{all} self-speculative decoding methods. Methods that operate on truly unmodified models, such as
$n$-gram speculation and Lookahead decoding either yield negligible speedups (since they only help on repetitive tasks) or lack support
across most models. Table~\ref{tab:training-comp} compares the training requirements of~\ours{} (for post-training modified EE models) against other competitive speculative-decoding methods.~\ours{}'s post-training adds fewer than 2\% additional parameters, completes in a few hours, and keeps the base model fully frozen. Moreover, EE-models are also readily available on HuggingFace, and our post-training procedure can convert any off-the-shelf model into an EE-model. Thus, even when the overhead of post-training is included, it remains comparable to or lower than the auxiliary training required by competitive speculative-decoding methods.

\begin{table}[htbp]
\centering
\caption{Training requirements of~\ours{} compared to competitive speculative-decoding methods. All methods require some form of auxiliary
training to achieve meaningful speedups.}
\label{tab:training-comp}
\setlength{\tabcolsep}{0pt}
\begin{tabular}{@{}lcccc@{}}
\toprule
& \textbf{EAGLE} & \textbf{Medusa} & \textbf{SPRINTER}     & \textbf{\ours{} \textit{(Ours)}} \\ \midrule
Trained component         & Transformer layer  & $K$ ResBlock heads   & FC classifier         & EE heads (LoRA) \\
Additional parameters     & $\sim$3\%          & $\sim$2--5\%         & $<$0.01\%             & $<$2\% \\
Base model modified       & No \cmark{}        & Varies \xmark{}      & No \cmark{}           & No \cmark{} \\
Training time             & hours--days        & hours--days          & hours                 & few hours \\
Pre-trained available     & Yes \cmark{}       & Limited \xmark{}     & No \xmark{}           & Yes (HF) \cmark{} \\
Lossless guarantee        & Yes \cmark{}       & Yes \cmark{}         & No \xmark{}           & Yes \cmark{} \\
\bottomrule
\end{tabular}
\end{table}

\section{Impact of Adaptive Hyper-parameter Selection}
\label{app:adapt-select}

To study the impact of dynamic hyper-parameter ($N_d$ and $N_i$) selection on~\ours{}, we design and implement an adaptive selection policy that adjusts these parameters in real-time based on the token acceptance rate at both the intermediate verification layer and the final model layer. For this study, we position the intermediate verification layer as shown in Table~\ref{tab:interim_layer}.

\subsection{Impact on Number of Draft Tokens per Step ($N_d$)}
\label{app:adapt-nd}

We design an adaptive policy to adjust $N_d$ based on the acceptance rate at the intermediate verification layer. This is motivated by the rationale that a high acceptance rate at the intermediate verification layer signals that the draft layer is generating high-quality draft tokens. Thus, a larger value of $N_d$ can be employed to improve overall throughput. Conversely, a low acceptance rate corresponds to low-quality draft tokens, indicating that a smaller $N_d$ would reduce the computational overhead of generating draft tokens that will likely be rejected. We set $N_i$ to 4 (default~\ours{}) for this study. We formally describe the policy used to adjust $N_d$ below:

Let $\alpha_{\text{inter}}^{(t)}$ denote the acceptance rate at the intermediate verification layer at step $t$. We define a high threshold ($\tau_{\text{high}}$ = 0.8) and low threshold ($\tau_{\text{low}}$ = 0.5) to either aggressively increase or conservatively reduce $N_d$. Furthermore, we bound this adaptive adjustment between $N_d^{\min}$ and $N_d^{\max}$ to prevent degenerate cases. We initialize $N_d^{\min}$ and $N_d^{\max}$ with 1 and 3 respectively (possible range bounded by $N_i$). The value of $N_d$ for the next step, $N_d^{(t+1)}$, is updated according to equation~\eqref{nd-rule}. Table~\ref{tab:nd_variation} shows that using an adaptive policy improves the throughput of~\ours{} by upto $2.1\%$.

\begin{equation}
N_d^{(t+1)} = \begin{cases}
\min(N_d^{\max}, N_d^{(t)} + 1) & \text{if } \alpha_{\text{inter}}^{(t)} \geq \tau_{\text{high}} \\
\max(N_d^{\min}, N_d^{(t)} - 1) & \text{if } \alpha_{\text{inter}}^{(t)} < \tau_{\text{low}} \\
N_d^{(t)} & \text{otherwise}
\end{cases}
\label{nd-rule}
\end{equation}

\begin{table}[h]
\centering
\vspace{-0.1in}
\caption{Throughput improvement of~\ours{} with a dynamic policy (initialized to $N_d = 2$) relative to the default ($N_d$ fixed at 2).\label{tab:nd_variation}}
\begin{tabular}{lcc}
\hline
\textbf{Model} & \textbf{ShareGPT} & \textbf{CNN/DM} \\
\hline
Llama2-13B$_{\textsc{inst}}$ & 1.22\% & 1.8\% \\
Llama2-70B$_{\textsc{inst}}$ & 2.1\% & 1.95\% \\
\hline
\end{tabular}
\vspace{-0.1in}
\end{table}

\subsection{Impact on Number of Tokens Tentatively Accepted ($N_i$)}
\label{app:adapt-ni}

Similarly, we design a policy to adjust $N_i$ based on the acceptance rate at the final layer. A high acceptance rate at the final layer indicates that the intermediate verification layer is accepting high-quality draft tokens which are likely to be accepted at the final layer. Thus, increasing $N_i$ would reduce the frequency of target (full-model) verification and improve throughput. Conversely, a low acceptance rate corresponds to low-quality draft tokens being accepted by the intermediate verification layer and necessitates more frequent target verification to avoid long-chains of unverified tokens which will likely be rejected by the final layer. We set $N_d$ to 2 (default~\ours{}) for this study. We formally describe the policy used to adjust $N_i$ below:

Let $\alpha_{\text{final}}^{(t)}$ denote the acceptance rate at the final layer at step $t$. We define a high threshold ($\tau_{\text{high}}$ = 0.8) and low threshold ($\tau_{\text{low}}$ = 0.3) to either aggressively increase or conservatively reduce $N_i$. We bound this adaptive adjustment between $N_i^{\min}$ and $N_i^{\max}$ to prevent long-chains of unverified tokens. We initialize $N_i^{\min}$ (bounded by $N_d$) and $N_i^{\max}$ to be 3 and 8, respectively. The value of $N_i$ for the next step ($N_i^{(t+1)}$) is updated using equation~\eqref{ni-rule}. Table~\ref{tab:ni_variation} shows that using an adaptive policy improves the throughput of~\ours{} by upto $4.2\%$.

\begin{equation}
N_i^{(t+1)} = \begin{cases}
\min(N_i^{\max}, N_i^{(t)} + 1) & \text{if } \alpha_{\text{final}}^{(t)} \geq \tau_{\text{high}} \\
\max(N_i^{\min}, N_i^{(t)} - 1) & \text{if } \alpha_{\text{final}}^{(t)} < \tau_{\text{low}} \\
N_i^{(t)} & \text{otherwise}
\end{cases}
\label{ni-rule}
\end{equation}

\begin{table}[h]
\centering
\vspace{-0.1in}
\caption{Throughput improvement of~\ours{} with a dynamic policy (initialized to $N_i = 4$) relative to the default ($N_i$ fixed at 4).\label{tab:ni_variation}\\}
\begin{tabular}{lcc}
\hline
\textbf{Model} & \textbf{ShareGPT} & \textbf{CNN/DM} \\
\hline
Llama2-13B$_{\textsc{inst}}$ & 3.0\% & 4.2\% \\
Llama2-70B$_{\textsc{inst}}$ & 3.7\% & 2.6\% \\
\hline
\end{tabular}
\vspace{-0.1in}
\end{table}

\section{Future Work}
\label{app:limitations}

\textbf{Integration with state-of-the-art draft methods:} The intermediate verification mechanism introduced in~\ours{} is orthogonal to the choice of the draft model. While we use an early-exit layer ($L_d$) for generating draft tokens,~\ours{} can be naturally extended to use other methods such as EAGLE~\cite{li2024eagle}, Medusa~\cite{cai2024medusa}, or any tree-based draft method. Tree-based draft methods verify several beams of tokens at once for each sequence, thus, the verification loads for these methods is very high. This makes~\ours{} uniquely suited to such approaches. We leave this integration as future work.

\textbf{Multiple intermediate verifiers:}~\ours{} currently uses a single intermediate verifier $L_i$ between $L_d$ and $L_f$. Extending this
to multiple intermediate verifiers at progressively deeper layers $L_d < L_{i,1} < L_{i,2} < \ldots < L_{i,k} < L_f$ could create a
finer-grained filtering hierarchy, where each verifier rejects increasingly subtle errors. The design space: selecting the number of
intermediate verifiers, their layer placements, and the per-stage tentative-acceptance windows ($N_{i,1}, \ldots, N_{i,k}$) is an open
direction, especially for very large models where additional intermediate stages could amortize the full-model verification over even more
tokens.

\end{document}